\documentclass{article}
\usepackage{arxiv}

\usepackage{graphicx}
\usepackage{amsmath}
\usepackage{amssymb}
\usepackage{comment}
\usepackage{color}
\usepackage{url}
\usepackage[normalem]{ulem}
\usepackage{acronym}
\usepackage{mathtools}
\usepackage{multirow}
\usepackage{multicol}

\definecolor{darkgreen}{RGB}{0,100,0}
\definecolor{orange}{RGB}{215,130,0}
\definecolor{grey}{RGB}{80,80,80}
\definecolor{navy}{RGB}{0,0,128}
\definecolor{purple}{RGB}{186,85,211}

\newcommand{\red}[1]{\textcolor{red}{#1}}
\newcommand{\best}[1]{\textbf{#1}}

\newtheorem{mydef}{Definition}

\newif\ifcomments
\commentstrue
\newcommand{\mycol}[2]{{\leavevmode\color{#1}#2}}

\newcommand{\commentIB}[1]{\ifcomments
\mycol{navy}{[\textbf{IB:} #1]}
\fi}

\newcommand{\commentST}[1]{\ifcomments
\mycol{orange}{[\textbf{ST:} #1]}
\fi}

\newcommand{\norm}[1]{\left\lVert#1\right\rVert}
\newcommand{\scnorm}[1]{\norm{#1}_2^s}
\newcommand{\prob}[2]{\mathbb{P}_{#1}\left(#2\right)}
\newcommand{\expectation}[2]{\mathbb{E}_{#1}#2}

\DeclareMathSymbol{\Gamma}{\mathalpha}{operators}{0}
\DeclareMathSymbol{\Delta}{\mathalpha}{operators}{1}

\DeclareMathOperator*{\minimize}{minimize}
\DeclareMathOperator*{\softmax}{softmax}
\newcommand\defop[1]{\mathrel{\stackrel{\makebox[0pt]{\mbox{\normalfont\tiny def}}}{#1}}}
\newcommand\defeq{\:\defop{=}\:}
\newcommand\defequiv{\:\defop{\Leftrightarrow}\:}

\newcommand{\titlestr}{Metrics and methods for robustness evaluation of neural networks with generative models}

\title{\titlestr}
\author{
  Igor Buzhinsky\\%\thanks{Use footnote for providing further
  Computer Technologies Laboratory, ITMO University, St. Petersburg, Russia, and\\
  Department of Electrical Engineering  and Automation, Aalto University, Espoo, Finland\\
  \texttt{igor.buzhinsky@gmail.com} \\
   \And
 Arseny Nerinovsky \\
 Computer Technologies Laboratory,\\ ITMO University, St. Petersburg, Russia \\
  \texttt{nerinovsky.arseny@gmail.com} \\
  \And
  Stavros Tripakis \\
  Northeastern University\\
  Boston, MA, US\\
  \texttt{stavros@northeastern.edu} \\
}

\begin{document}
\maketitle

\begin{abstract}
Recent studies have shown that modern deep neural network classifiers are easy to fool, assuming that an adversary is able to slightly modify their inputs.
%For example, in image classification, a misclassification may often be achieved even with a human-imperceptible change of pixels.
Many papers have proposed adversarial attacks, defenses and methods to measure robustness to such adversarial perturbations.
However, most commonly considered adversarial examples are based on $\ell_p$-bounded perturbations in the input space of the neural network, which are unlikely to arise naturally.
Recently, especially in computer vision, researchers discovered ``natural'' or ``semantic'' perturbations, such as rotations, changes of brightness, or more high-level changes, but these perturbations have not yet been systematically utilized to measure the performance of classifiers.
In this paper, we propose several metrics to measure robustness of classifiers to natural adversarial examples, and methods to evaluate them.
These metrics, called latent space performance metrics, are based on the ability of generative models to capture probability distributions, and are defined in their latent spaces.
%Recent studies have shown that modern deep neural network classifiers are often easy to fool, assuming that an adversary is able to slightly modify or choose their inputs.
%For example, in image classification, a misclassification may often be achieved even with a human-imperceptible change of pixels.
%While the first adversarial examples were based on $\ell_p$-bounded perturbations in the input space of a neural network, which are unlikely to arise naturally, later researchers discovered more ``natural'' or ``semantic'' perturbations, such as corruption with noise, rotations, changes of brightness or changes in the latent space of a generative model.
%Exploiting the fact that generative models are models of distributions, this paper studies various ways to specify and evaluate performance metrics for neural network classifiers with their help, which we call latent space performance metrics.
%As a result, classifier evaluation is \red{enhanced with the priors coming from a concrete classification problem} \commentIB{I think that I cannot elaborate. Maybe instead we should return to the talk about safety concerns and an adversary that manipulates with semantic features if the classified object.}.
On three image classification case studies, we evaluate the proposed metrics for several classifiers, including ones trained in conventional and robust ways.
We find that the latent counterparts of adversarial robustness are associated with the accuracy of the classifier rather than its conventional adversarial robustness, but the latter is still reflected on the properties of found latent perturbations.
In addition, our novel method of finding latent adversarial perturbations demonstrates that these perturbations are often perceptually small.

\end{abstract}

%\commentST{Stavros' comments on the abstract, after discussing with Igor during 14 feb meeting at Aalto: the ``story'' on the abstract should be:\\
%(1) I (Igor) don't like the current metrics that people use for robustness: why? because they operate not on the latent space, which means they ``mostly'' (?) cover unnatural counter-examples;\\
%(2) So I propose new metrics, which operate on the latent space: the latent space is good, because it covers mostly natural counter-examples;\\
%(3) I'm using GANs and autoencoders to capture the latent space and the metrics;\\
%(4) Here are my results.}

%\commentIB{Additional comments:\\
%-- People came up with attacks and defenses. I come up with metrics.\\
%-- Methods should be somehow addressed in addition to metrics. This can be mentioned in the abstract as a co-contribution.\\
%-- Missing part to ``why generative models'': in the latent space, all examples are in some sense natural, unlike in the original space.
%-- Papers on latent adversarial examples and generative adversarial examples can be used to motivate metrics based on reconstruction and generation.
%-- Adversarial severity and frequency should be introduced in Preliminaries (existing metrics).\\
%-- In LLAR, we look at the norm of what? Of a difference between the decayed vector and the perturbed vector. This should be clarified.\\
%-- Exclude ``adversarial'' from the title. ``Metrics and methods for evaluating robustness of neural networks in latent spaces''?\\
%}

\section{Introduction}
\label{sec:intro}

Unlike in more conventional software engineering, the problem of ensuring reliability of machine learning (ML) based software is complicated by the fact that ML-based models, such as \emph{artificial neural networks} (ANNs), are not programmed explicitly.
Instead, they significantly depend on the data on which they are trained.
The traditional form of assessing model performance based on validation/test data (e.g., a holdout set) and measures such as accuracy or F-score, become insufficient when the models interact with the real world, such as in the cases of aircraft and unmanned vehicle control.
%This is proven by the discovery of \emph{adversarial examples}~\cite{szegedy2014intriguing}---small input changes that cause ML models to make mistakes, and by a volume of evidence~\cite{gilmer2018motivating,akhtar2018threat} that adversarial examples are not only a cyber-phenomenon but they transfer to the material world.
This is proven by the discovery of \emph{adversarial examples}~\cite{szegedy2014intriguing}---slightly perturbed inputs that cause ANNs to malfunction, for example by misclassifying an image.
For a human, adversarial examples may be even indistinguishable from original, unperturbed inputs.
Adversarial examples are often produced in a rather artificial environment, by adopting special algorithms that perturb the input until a certain criterion is reached, but recent evidence~\cite{akhtar2018threat,gilmer2018motivating} suggests that they may transfer to the material world.
The classic framework of \emph{empirical risk minimization} (ERM)~\cite{vapnik2013nature}, where the classifier is trained on available data samples, is used to achieve high values of sample-based metrics such as accuracy or F-score.
%If the samples are drawn from the real data distributions, then accuracy becomes an unbiased estimate of the classifier's performance on unknown samples, and it is possible to speak of a constraint on accuracy or a similar metric as a form of specification.
However, if the classifier must be protected from adversarial examples, ERM is insufficient, and robust optimization~\cite{madry2017towards} with projected gradient descent (PGD) can be used instead.
This corresponds to enforcing \emph{adversarial robustness}~\cite{anderson2019optimization,bastani2016measuring,fawzi2018analysis,huang2017safety,katz2017reluplex,moosavi2016deepfool,singh2019abstract}, which is often treated either as a metric~\cite{bastani2016measuring,fawzi2018analysis,moosavi2016deepfool} specifying the minimum magnitude of an adversarial perturbation, or
%, in the context of formal verification,
as a specification~\cite{anderson2019optimization} stating that the decision of the ANN must be invariant to perturbations of input of a certain form.
Adversarial robustness can be local~\cite{anderson2019optimization,fawzi2018analysis,huang2017safety,katz2017reluplex,singh2019abstract} (for a particular input) or global~\cite{katz2017reluplex} (for all inputs).

Traditional adversarial examples are based on perturbations in the input space of the ANN that are constrained with $\ell_p$ (e.g., $
\ell_2$ or $\ell_\infty$) norms.
The resulting adversarial examples are highly improbable to arise naturally~\cite{song2018pixeldefend}, but it was shown that even \emph{natural adversarial examples} (i.e., the ones plausible under the data distribution) exist~\cite{amadou2019semantics,dreossi2018semantic,engstrom2019exploring,gu2019using,hendrycks2019natural,ilyas2019robust,song2018constructing,zhao2017generating}.
While conventional adversarial examples often require 2D or 3D printing of precomputed images~\cite{akhtar2018threat} to be applied in the real world, a natural adversary could be seen as a manipulator of semantic features of classified objects.
Construction of a subclass~\cite{amadou2019semantics,ilyas2019robust,song2018constructing,zhao2017generating} of such examples is possible with the help of generative models, such as generative adversarial networks (GANs)~\cite{goodfellow2014generative} and generative autoencoders~\cite{makhzani2015adversarial}.
Previous works that considered natural adversarial examples mostly focused on attacks (e.g.,~\cite{ilyas2019robust,zhao2017generating}) and defenses (e.g.,~\cite{ilyas2019robust,samangouei2018defense,song2018pixeldefend}) rather than assessing the performance of classifiers.
Natural adversarial examples were also applied for ANN training~\cite{ilyas2019robust}, although the focus so far has been on adversarial robustness in the input space of the ANN.

%Other works enforced the output of an ML model to be within a required range~\cite{pulina2010abstraction,dutta2017output,ruan2018reachability}, ensured the fairness of ML models~\cite{zemel2013learning}, or proposed to augment ML models with safety \emph{shields}~\cite{bloem2015shield,alshiekh2018safe}.

%\commentIB{The connection of the paper to deep learning must be shown clearly!}

%\commentST{this should go in the introduction section.}

This paper utilizes generative models as a means of capturing real-world data distributions in order to move closer to formalizing and checking the original, natural language specification: \emph{``the ANN shall classify the input to class $c$ whenever the input belongs to class $c$.''}
%formalizing, checking and enforcing
Towards this end, our paper studies how to specify and evaluate \emph{performance metrics} for ANN classifiers in terms of probabilities, likelihood and distances in latent spaces of generative models.
As a result, our metrics capture the robustness of classifiers to natural adversarial examples, and thresholds on these metrics may correspond to the desired specifications.
%The possibility of using such models to perform search in their latent spaces opens new prospects to formalize, check and, potentially, enforce specifications in terms of probabilities, likelihood and distances in these spaces.
%The most general problem that we aim to address in this paper is the problem of \emph{formulating and checking high-level specifications for feed-forward ANN classifiers}.
%So far, checkable formal specifications for ANNs differ a lot from the original, natural language~(NL) specification: \emph{``the ANN shall classify the input to class $c$ whenever the input belongs to class $c$.''}
%How can this difficult-to-formalize NL specification be combined with generalization beyond the distribution $\mathcal{D}$ that is given in training samples?
%While not proposing a way to formalize this specification precisely, we attempt to move in this direction by proposing a model-based approach that utilizes an approximation of $\mathcal{D}$ with a latent distribution to consider possible inputs not used during training as well as the plausibility of these inputs.
%This enables probabilistic reasoning and a ``soft'', probabilistic formulation of specifications.
The contributions of the paper are as follows:
\begin{enumerate}
\item We propose a framework to evaluate the performance of feed-forward deep ANN classifiers with the help of generative models and their latent spaces.
The implementation of the framework is publicly available online.
\item Within this framework, we propose \emph{latent space performance metrics}---novel performance metrics for feed-forward ANN classifiers that are grounded on probabilistic reasoning in latent spaces of generative models, and, informally speaking, measure the ``resistance'' of the classifier to natural adversarial examples~\cite{ilyas2019robust,song2018constructing,zhao2017generating}.
The naturality of adversarial examples is achieved by (1) operating in the latent space of the generative model, (2) considering a distribution-preserving model of noise, and (3) generating adversarial examples by adding random noise, or by searching for worst-case examples that are bounded by the likelihood of noise.
\item We propose methods to approximately evaluate these metrics in a white-box setting using (1) sampling and (2) gradient-based search of adversarial perturbations in the latent space.
The latter method is a form of untargeted attack based on PGD.
We show that such a search is possible not only with GANs~\cite{zhao2017generating}, but also with generative autoencoders.
\item For each of three considered image classification case studies, we examine five classifiers trained traditionally and in a way that achieves adversarial robustness, and evaluate their performance according to latent space performance metrics.
Our PGD-based untargeted attack is able to find perceptually smaller latent perturbations than reported earlier~\cite{zhao2017generating}, and we find positive association between latent counterparts of adversarial robustness and the accuracy of a classifier on clean images.
What is more, this association is absent for latent space performance metrics and conventional adversarial robustness, but instead the latter leads to minimum latent adversarial perturbations being further from the original image in the original (non-latent) space as well as perceptually.
%\item \red{We propose a new method of robust optimization that is targeted on enforcing one of the aforementioned specifications.
%This method ensures robustness both in the latent space and in the input space of the ANN classifier.} \commentIB{Not yet done. May require MNIST because working with CelebA is too computationally hard. Maybe this should be left out of this paper.}
\end{enumerate}

The rest of the paper is structured as follows.
Section~\ref{sec:preliminaries} presents background material.
Section~\ref{sec:metrics} motivates the use of generative models to measure ANN classifier performance, and proposes corresponding metrics.
In Section~\ref{sec:measuring_metrics}, approaches are given to evaluate these metrics.
Evaluation of deep convolutional neural network (CNN) classifiers with these approaches is performed in Section~\ref{sec:experiments}.
Section~\ref{sec:related} reviews related work, and Section~\ref{sec:conclusion} concludes the paper.

\section{Preliminaries}
\label{sec:preliminaries}

\subsection{Artificial neural networks}

% AN_EDIT: change ... (ANN) is a parametric model to predict...
% AN_EDIT: to ... (ANN) a parametric model that  predicts...
A \emph{feed-forward artificial neural network} (ANN) $\mathcal{N}$ is a parametric model that predicts some outcome $y$ (a single number or a vector) based on some input vector $x$ of dimension $n_I$.
By feed-forward, we mean that the input is supplied to the network at once and is passed through a predefined computation graph with a finite number of computation nodes.
When the input is an image, $\mathcal{N}$ is usually a \emph{convolutional neural network} (CNN).
In this paper, we focus on the \emph{classification task}, where $\mathcal{N}$ must assign its input to one of $m > 1$ classes.
Thus, we have $\mathcal{N}: \mathbb{R}^{n_I} \to \{1, ..., m\}$.
We assume that class prediction is done as follows: $\mathcal{N}$ first produces real-valued scores of each class $i$, to which we will refer as the values of the \emph{scoring function} $S_\mathcal{N}(x, i)$, and the actually predicted class is the one with the maximum score: $\mathcal{N}(x) = \arg\max_i S_\mathcal{N}(x, i)$.
In addition, we require that $S_\mathcal{N}(x, i)$ is continuous and almost everywhere differentiable with respect to $x$.

%\AN{$\text{V1:}$ We apply the following notation $\mathcal{N}(x) = \arg\max_i [\mathcal{N}'(x)]_i$ where $[\mathcal{N}'(x)]_i$ is the $i$-th element of the $\mathcal{N}'(x)$ probability vector.}
%\AN{$\text{V2:}$ use $S$ instead of  $\mathcal{N}'$}
%\commentIB{Problems with this approach: (1) $\mathcal{N}'$ is not yet defined, (2) square bracket notation is already used for a different purpose in this paper, (3) values before argmax are not required to be probability vectors. How about using notation $S_\mathcal{N}(x, i)$ and mentioning that it is actually the network itself, but without the last decision layer?}

%\commentST{maybe some citation here, for CNN or for NN in general, for entire paragraph.} \commentIB{These are too well-known things to cite.}
ANN classifiers are typically trained in a supervised way with some form of \emph{gradient descent} (e.g., stochastic gradient descent), using samples $x_1, ..., x_k \in \mathbb{R}^{n_I}$, which are paired with respective reference class labels $y_1, ..., y_k \in \{1, ..., m\}$.
These pairs $(x_1, y_1), ..., (x_k, y_k)$ are assumed to come from \emph{joint distribution} $\mathcal{D}^\text{joint}$, whose marginals are \emph{input data distribution} $\mathcal{D}$ and the \emph{class label distribution} $\mathcal{D}^\text{labels}$.

\subsection{Generative models}
\label{sec:gm}

A \emph{generative adversarial network} (GAN)~\cite{goodfellow2014generative}, which consists of two feed-forward ANNs called the \emph{discriminator} and the \emph{generator} $\mathcal{G}$, is trained to make $\mathcal{G}$ generate elements of some target data distribution $\mathcal{D}$ of $n_I$-dimensional vectors (in the simplest case, without sample labels).
Data generation is done by applying $\mathcal{G}$ to a low-dimensional vector $l \in \mathbb{R}^{n_L}$ sampled from the \emph{latent code distribution} $\mathcal{D}_L$ (typically, $N(0, I)$).
If $l \sim \mathcal{D}_L$, then for a well-trained GAN we may assume that $\mathcal{G}(l) \sim \mathcal{D}$.
Often, the dimension of $\mathcal{D}_L$ is made smaller than the dimension of $\mathcal{D}$: $n_L < n_I$.
The set of all latent codes (usually, just $\mathbb{R}^{n_L}$) is called the \emph{latent space}.
By contrast, we will refer to the input space of an ANN classifier ($\mathbb{R}^{n_I}$) as the \emph{original space}.
With some enhancements, GANs may be also capable of \emph{reconstruction}---finding latent representation $l \in \mathbb{R}^{n_L}$ for the given original vector $x \in \mathbb{R}^{n_I}$ such that $\mathcal{G}(l)$ is close to $x$ (e.g., according to some norm in the original space).
For example, this may be done by training an additional ANN $\mathcal{I}: \mathbb{R}^{n_I} \to \mathbb{R}^{n_L}$ called an \emph{inverter}~\cite{hendrycks2019natural}.
However, obtaining good inversions, especially for GANs that generate high-resolution images, requires more effort: for example, in~\cite{bau2019seeing}, inversion is performed layer-wise and combined with gradient-based optimization.

An \emph{autoencoder} $(\mathcal{N}^E, \mathcal{N}^D)$, where $\mathcal{N}^E$ and $\mathcal{N}^D$ are feed-forward ANNs called the \emph{encoder} and the \emph{decoder} respectively, is a model whose goal is to \emph{compress} (\emph{encode}) its inputs $x \in \mathbb{R}^{n_I}$ to low-dimensional vectors $l = \mathcal{N}^E\left(x\right) \in \mathbb{R}^{n_L}$ (again, $n_L < n_I$) such that approximate \emph{decompression} (\emph{decoding}, \emph{reconstruction}) can be achieved: $\mathcal{N}^D\left(l\right)$ is close to $x$.
%e.g., such that the \emph{reconstruction error} $\operatorname{E}_{x \sim D}{\norm{x - \mathcal{N}^D(\mathcal{N}^E(x))}}$ is small according to some norm $\norm{\cdot}$ (for example, $\norm{\cdot} = \norm{\cdot}_2$).
%\commentIB{\soutIB{What's written above doesn't allow randomized choices, e.g., sampling from the distribution produced by the encoder (as in variational autoencoders). Maybe this should be added to the definition.} The approach is only applicable to deterministic autoencoders. No need to mention VAEs.}
A \emph{generative autoencoder} (such as in~\cite{heljakka2020towards,makhzani2015adversarial}) is an autoencoder whose decoder is additionally trained to sample from the original distribution $\mathcal{D}$---thus, essentially, a generative autoencoder performs both the tasks of an autoencoder and a GAN.
%: $y = \mathcal{N}_D(\mathcal{N}_E(x))$ must be made plausible according to the original data distribution $\mathcal{D}$.
For a well-trained generative autoencoder, we may assume both $l \sim \mathcal{D}_L \Rightarrow \mathcal{N}^D(l) \sim \mathcal{D}$ and $x \sim \mathcal{D} \Rightarrow \mathcal{N}^E(x) \sim \mathcal{D}_L$.
%However, in practice, if generation is done from distribution $\mathcal{D}_L$ chosen beforehand, such as $N(0, I)$, the distribution of the encoder's output may be slightly different.
%Formally, consider the \emph{latent code distribution} $\mathcal{D}_L$ such that $\mathcal{N}_E(x) \sim \mathcal{D}_L$ when $x \sim \mathcal{D}$.
%Then if $l \sim \mathcal{D}_L$, the decoder produces $y = \mathcal{N}_D(l) \sim \mathcal{D}'$, where $\mathcal{D}'$ is similar to $\mathcal{D}$ (ideally, $\mathcal{D}' = \mathcal{D}$).
%In this case, the decoder behaves like a generator of a generative adversarial network (GAN), and the aforementioned similarity of $\mathcal{D}'$ and $\mathcal{D}$ can be defined as the inability of the discriminator ANN (or the encoder~\cite{heljakka2018pioneer}) to distinguish them.

To summarize, generative models are capable of data \emph{generation} from low-dimensional vectors.
By using special types of generative models or enhancing existing generative models, it is also possible to achieve data \emph{reconstruction}.

\subsection{Adversarial examples and perturbations}

%\commentAN{Probably a somewhat different wording would be better here
%
%An \emph{adversarial example} is an input to an ANN classifier $\mathcal{N}$, $x'$ such that  $\norm{x' - x}_p \le \epsilon$ and $\mathcal{N}(x') \ne \mathcal{N}(x)$, where $x$ is a real data sample, $\norm{\cdot}_p$ is an $l_p$ norm, usually $l_2$ or $l_\inf$ norms are used.  $\Delta x = x' - x$ is usually referred to as an \emph{adversarial perturbation}.
%
%% Adversarial examples are usually defined, in terms of  a $l_p$ norm and a bound on possible perturbation $\epsilon$
%}

Suppose that $\mathcal{N}$ is an ANN classifier.
An \emph{adversarial example} is an input $x'$ to $\mathcal{N}$ such that $x' \in A(x)$ and $\mathcal{N}(x') \ne \mathcal{N}(x)$, where $x$ is a real data sample, $A(x)$ is the set of allowed changes of $x$ (often, it is taken as the $\varepsilon$-ball around $x$ according to the $\ell_p$ norm: $A(x) = \{x' \: | \norm{x' - x}_p \le \varepsilon\}$).
$\Delta x = x' - x$ is the corresponding \emph{adversarial perturbation}.

%Suppose that $\mathcal{N}$ is an ANN classifier.
%Usually, for some norm $\norm{\cdot}$ and a bound $\epsilon$ on it, an \emph{adversarial example} is defined as a change $x'$ of a real data sample $x$, such that $\norm{x' - x} \le \epsilon$ and $\mathcal{N}(x') \ne \mathcal{N}(x)$.
%In this context, $\Delta x = x' - x$ is the corresponding \emph{adversarial perturbation}.
%\commentAN{Added some blank lines here}

Adversarial examples and adversarial perturbations have been first found to exist in~\cite{dalvi2004adversarial,globerson2006nightmare}, but became more known from the work~\cite{szegedy2014intriguing}, where human-indistinguishable ImageNet perturbations were presented.
Since 2013, many~\cite{akhtar2018threat} adversarial attacks and defenses have been proposed.
While many proposed defenses were shown to be ineffective~\cite{gilmer2018motivating}, attacks were transported to the real world~\cite{akhtar2018threat}, raising concerns regarding the safety and security of deep ANNs.

%\commentAN{
%Another version of the paragraph below
%
%For adversarial perturbations bounded with $\ell_2$ and $\ell_\infty$ norms, \emph{projected gradient descent} (PGD) has been shown~\cite{madry2017towards} to be the best adversary that has access only to $\nabla_x S_\mathcal{N}(x, \cdot)$.
%The most common and most effective method of defense is \emph{robust optimization}, where adversarial examples for the current version of the ANN are injected in the training dataset. In~\cite{ford2019adversarial}, it was shown that it is possible to train the classifier on samples with added visible Gaussian noise instead of specially crafted adversarial examples.
%
%}

For adversarial perturbations bounded with $\ell_2$ and $\ell_\infty$ norms, \emph{projected gradient descent} (PGD) has been shown~\cite{madry2017towards} to be the best adversary that has access only to $\nabla_x S_\mathcal{N}(x, \cdot)$.
The most common
%\soutIB{and most effective} \commentIB{There are other methods. Training on noise images is a bit worse, but I hesitate to draw attention to this. There is also a certified defense method which is claimed to be as effective as robust optimization.} \commentAN{Ok}
method of defense is \emph{robust optimization} with PGD, where training is done on adversarial examples for the current version of the ANN.
In~\cite{ford2019adversarial}, it was shown that it is possible to train the classifier on samples with added visible Gaussian noise instead of specially crafted adversarial examples.

%Yet, for adversarial perturbations bounded with $\ell_2$ and $\ell_\infty$ norms, \emph{projected gradient descent} (PGD) has been shown~\cite{madry2017towards} to be the best adversary that has access only to $\nabla_x S_\mathcal{N}(x, \cdot)$.
%The corresponding method of defense is \emph{robust optimization} with PGD, where training is done on adversarial examples for the current version of the ANN.
%In~\cite{ford2019adversarial}, it was shown that instead it is possible to train the classifier on samples with added visible Gaussian noise.

Recent works explain adversarial examples through the peculiarities of the multidimensional geometry~\cite{ford2019adversarial} and the fact that conventional ERM-based training does not introduce human priors to the training process~\cite{ilyas2019adversarial}.
It has been also hypothesized~\cite{samangouei2018defense,song2018pixeldefend} that adversarial examples do not lie on the data manifold of the training distribution, but several works show that even \emph{natural} adversarial examples exist, such as the ones that come from the real world~\cite{hendrycks2019natural}, are made by rotations and translations~\cite{engstrom2019exploring}, color distortions~\cite{gu2019using}, semantic changes~\cite{dreossi2018semantic}, looping over consequent video frames~\cite{gu2019using}, and created with generative models~\cite{amadou2019semantics,ilyas2019robust,song2018constructing,zhao2017generating}.
\emph{Latent space adversarial examples}, or adversarial examples that correspond to some latent codes of a generative model, may be based on perturbations~\cite{amadou2019semantics,zhao2017generating} or generated from scratch~\cite{song2018constructing}.
It was shown~\cite{ilyas2019robust} that latent space adversarial examples can be used to enhance robust optimization and increase the overall robustness of the classifier.

\subsection{Performance metrics for adversarial robustness}
\label{sec:usual_perf_metrics}

Often, the set of possible adversarial examples is defined locally for each input $x$---for example, as an $\ell_p$ $\varepsilon$-ball, or as a set of rotations of $x$~\cite{engstrom2019exploring}.
The robustness of the classifier is then measured as its accuracy on worst-case inputs taken from such sets.
In~\cite{bastani2016measuring}, for the $\ell_\infty$ norm, this metric was formalized as \emph{adversarial frequency}.
Adversarial frequency, however, depends on $\varepsilon$.
A different way to measure robustness, which is free from this hyperparameter, is \emph{adversarial severity}~\cite{bastani2016measuring}---the expected (with $x \sim \mathcal{D}$) minimum distance from $x$ to an adversarial example.
The corresponding local metric is \emph{pointwise robustness}, which is the minimum distance to an adversarial example for a particular $x$.
In this paper, we will define metrics that are based on pointwise robustness, adversarial frequency and severity but operate with different norms in different spaces.
Known metrics that are defined in the original space will be referred to as \emph{conventional} metrics.

\section{Latent space performance metrics}
\label{sec:metrics}

In this paper, we are interested in \emph{specifying and evaluating performance metrics for ANN classifiers with the help of generative models}.
In addition, we would like to evaluate these metrics given the original training and validation data.
This section will propose several such \emph{latent space performance metrics}, and methods to evaluate them will be proposed in Section~\ref{sec:measuring_metrics}.

%\commentAN{approximately? this part is somewhat confusing, why should we evaluate them approximately ?}

\subsection{Preliminary definitions}

Suppose that $\mathcal{N}: \mathbb{R}^{n_I} \to \{1, ..., m\}$, $m > 1$, is a feed-forward ANN classifier with scoring function $S_\mathcal{N}$.
The goal of $\mathcal{N}$ is to correctly classify input vectors drawn from distribution $\mathcal{D}$.
In the most general case, there may be no unique correct label for an input vector, but rather there is a \emph{joint distribution} $\mathcal{D}^\text{joint}$ of pairs $(x, y)$ of an input vector $x$ and its label $y$.
For simplicity, we assume that $\mathcal{N}$ is validated on samples drawn exactly from $\mathcal{D}^\text{joint}$, although the training might have been performed on a distribution induced by data augmentation of input vectors $x$.
%Due to data augmentation, $\mathcal{D}$ may not be the distribution on which $\mathcal{N}$ is trained: such possibly augmented distribution we will denote as $\mathcal{D}_\text{train}$

Suppose that $\mathcal{D}_L^{i}$, $1 \le i \le m$, are $n_L$-dimensional ($n_L < n_I$) \emph{class-conditional latent distributions} (often assumed to be $N(0, I)$) such that we have trained transformations $D_i$ that generate samples from class-conditional data distributions $\mathcal{D}^{i}$: $l \sim \mathcal{D}_L^{i} \Rightarrow D_i(l) \sim \mathcal{D}^{i}$.
In certain cases (see models capable of reconstruction in Section~\ref{sec:gm}), we may additionally have transformations $E_i$ that return latent code approximations of $n_I$-dimensional vectors.
We would like $D_i$ to be compatible with %backpropagation,
gradient descent,
i.e., continuous and almost everywhere differentiable, but we do not require the same from $E_i$.

\subsection{Motivation for latent space performance metrics}
\label{sec:problem}

With both $D_i$ and $E_i$, we can convert vectors to the latent space and back.
Assuming that the latent space corresponds to a well-trained generative model, working in it has the following benefits compared to the original space:
\begin{enumerate}
\item For a random latent vector $l \sim \mathcal{D}_L^{i}$, $D_i(l)$ is often plausible according to $\mathcal{D}$.
\item Changes of the vector in the latent space are \emph{semantic} (high-level) in terms of the original representation.
\item For each class $i$, the image $D_i(\mathbb{R}^{n_L})$ contains an infinite number of diverse data samples, which may be useful for evaluation and further training of $\mathcal{N}$.
\item The aforementioned samples can not only be generated at random, but also can be optimized with gradient-based techniques to optimize a certain objective (e.g., $S_\mathcal{N}$).
\end{enumerate}

As many performance metrics, such as accuracy, adversarial frequency and severity, will remain meaningful when the original space is replaced with the latent one, the main approach of introducing latent performance metrics used in this paper is \emph{moving conventional performance metrics to the latent space}.
We will do it in a way that provides additional benefits related to the probabilistic interpretation of the latent space---for example, while considering adversarial perturbations, we will take care that the data remains plausible according to $\mathcal{D}$.

\subsection{Possible scenarios}

\begin{figure}
\centering
\includegraphics[width=0.9\textwidth]{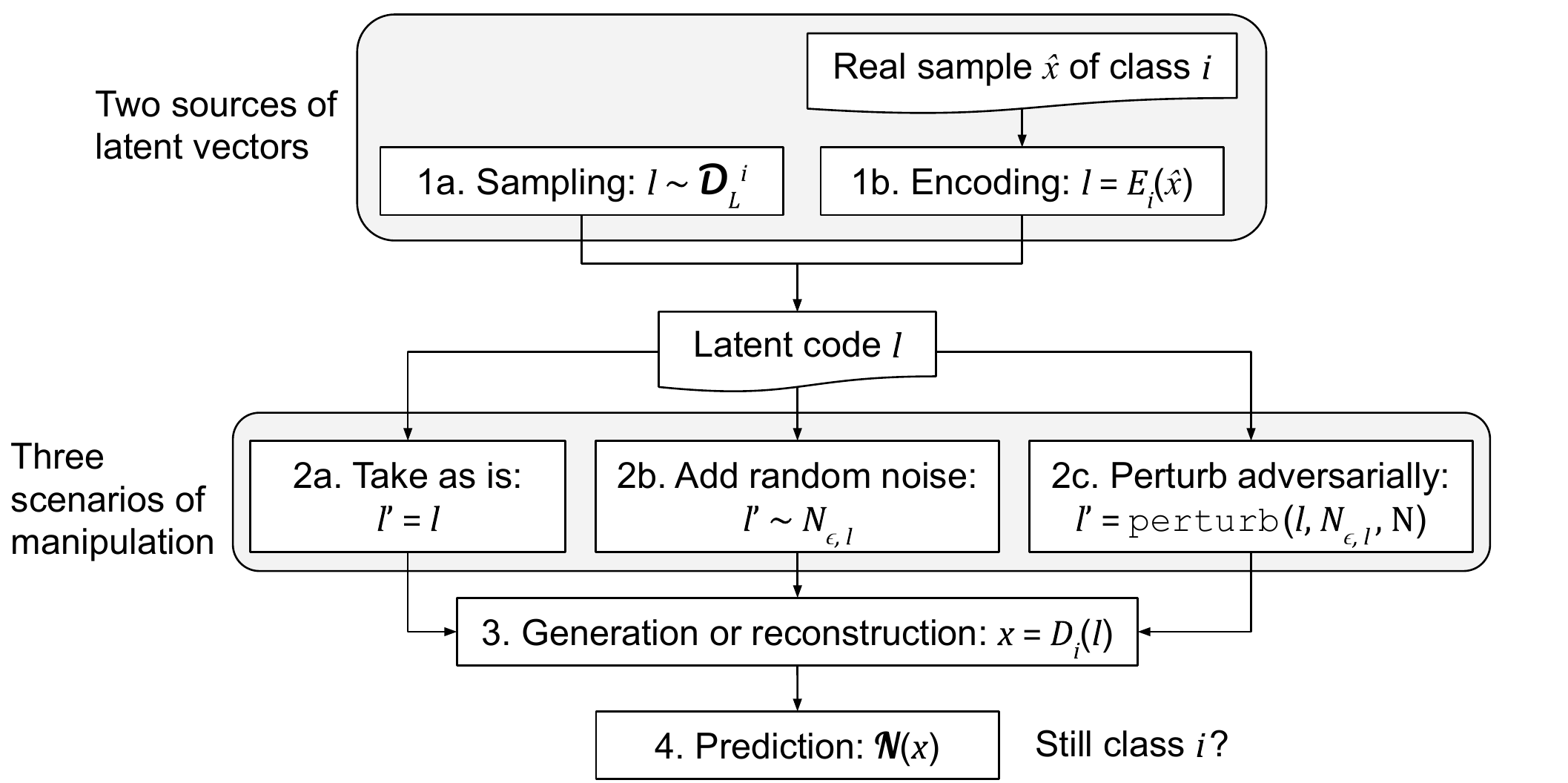}
\caption{Overview of considered scenarios.}
\label{fig:problem_overview}
\end{figure}

Intuitively, sampling from $\mathcal{D}_L^{i}$ gives latent vectors $l$ such that $D_i(l)$ are instances of class $i$.
Previous works on natural adversarial examples obtained latent vectors based on generation~\cite{song2018constructing} and reconstruction~\cite{zhao2017generating}.
These two scenarios of obtaining $D_i(l)$ (see the upper part of Fig.~\ref{fig:problem_overview}, paths 1a and 1b) directly correspond to two operations that generative models are capable of (see Section~\ref{sec:usual_perf_metrics}):
%Consider two scenarios:
\begin{enumerate}
\item sample $l \sim \mathcal{D}_L^{i}$ and \emph{generate} $x = D_i(l)$;
\item take a random real sample $\hat{x} ~ \sim \mathcal{D}^{i}$, encode it as $l = E_i(\hat{x})$, and \emph{reconstruct} it as $x = D_i(l) $.%, pretending that it has been sampled from $\mathcal{D}_L^{1}$.
%Suppose that there is a randomized noise adding transformation $N_{\epsilon}^{c=i}$ such that if $l \sim \mathcal{D}_L^{i}$, then $N_{\epsilon}^{c=i}(l) \sim \mathcal{D}_L^{i}$ (the distribution is preserved), where parameter $\epsilon$ controls the magnitude of the noise.
%Would $N_{\epsilon}^{c=i}(l_0)$ be still classified as a kitten?
%With what probability can this be achieved for different values of $\epsilon$?
%Are there noise additions with high likelihood that can cause $N_{\epsilon}^{c=i}(l_0)$ to be misclassified?
\end{enumerate}

In this paper, we are interested in finding latent space counterparts for the following metrics (each of them will correspond to one of three scenarios in the lower part of Fig.~\ref{fig:problem_overview}):
\begin{enumerate}
\item \textbf{Accuracy}, as well as similar metrics based on counting success frequencies (Fig.~\ref{fig:problem_overview}, path 2a).
This is the simplest case: it is sufficient to calculate the success frequency of $\mathcal{N}$ on reconstructed or generated samples.
This will be formalized in Section~\ref{sec:gla}.
\item \textbf{Corruption robustness to random noise}~\cite{ford2019adversarial,hendrycks2019benchmarking} (Fig.~\ref{fig:problem_overview}, path 2b).
While in the original space the addition of noise is a form of data corruption, in the latent space this noise will introduce semantic changes to the input, and we could measure the success frequency of $\mathcal{N}$ on such semantically modified inputs.
In Section~\ref{sec:noise_accuracy}, we will introduce a family of noise-adding distributions $N_{\epsilon, l}$ that retain the transformed data plausible even for large noise, and define a corresponding performance metric.
\item \textbf{Adversarial robustness}~\cite{anderson2019optimization,fawzi2018analysis,huang2017safety,katz2017reluplex,singh2019abstract} (Fig.~\ref{fig:problem_overview}, path 2c).
Adversarial robustness in the latent space can be treated as ``resistance'' to worst-case noise additions that are bounded according to noise likelihood and optimized to degrade the performance of $\mathcal{N}$.
The connection between noise corruption robustness and adversarial robustness exists already in the original space: for example, if the noise is Gaussian, its likelihood is determined by its $\ell_2$ norm, a threshold on which is a common constraint on adversarial perturbations.
What is more, noise corruption robustness and adversarial robustness were found to be highly related~\cite{ford2019adversarial}.
The corresponding latent space metrics will be formalized in Section~\ref{sec:latent_robustness}.
\end{enumerate}

%\paragraph{Scenario 2:} Would $\mathcal{N}$ be robust to small changes $l'$ of $l$ sampled from $N_{\epsilon, l}$?
%\paragraph{Scenario 3:} Would $\mathcal{N}$ be robust to small changes $l'$ of $l$ that are crafted by an adversary and are bounded with the likelihood according to $N_{\epsilon, l}$?

%\commentST{i would expect at the end of this section a formal problem definition, of the form: Given X Find Y such that Z. in the absence of such a definition, it's hard to call this section ``problem formulation''}

%Given an ANN classifier $\mathcal{N}$, class-conditional latent code distributions $\mathcal{D}_L^{i}$, a family of noise-adding distributions $N_{\epsilon, l}$, class-conditional generative models $D_i$ and, optionally, latent code reconstruction transformations $E_i$, we aim to formalize the questions above
%in performance metrics of $\mathcal{N}$ that, in the general case, can be formalized as follows:
%\begin{itemize}
%\item a \emph{global performance metric} is a real-valued function $M(\mathcal{N}, S, \mathcal{D}_L^{1}, ..., \mathcal{D}_L^{m}, N_{\epsilon, l}, D_1, ..., D_m, E_1, ..., E_m)$;
%\item a \emph{local performance metric} also accepts a real data instance $(x, i)$.
%\end{itemize}

\subsection{Accuracy in the latent space}
\label{sec:gla}

Probably the simplest thing that can be done with generative models is to evaluate the accuracy of the classifier on generated and reconstructed data items.
This situation corresponds to the absence of any adversary.
These ideas are formalized in the following definitions:

\begin{mydef}[latent generation accuracy]
the latent generation accuracy (LGA) of $\mathcal{N}$ is:
$$\mathrm{LGA}(\mathcal{N}) \defeq \prob{i \sim \mathcal{D}^\text{labels},\; l \sim \mathcal{D}_L^{i}}{\mathcal{N}(D_i(l)) = i}.$$
\end{mydef}

\begin{mydef}[latent reconstruction accuracy]
the latent reconstruction accuracy (LRA) of $\mathcal{N}$ is:
$$\mathrm{LRA}(\mathcal{N}) \defeq \prob{(x, i) \sim \mathcal{D}^\text{joint}}{\mathcal{N}(D_i(E_i(x))) = i}.$$
\end{mydef}

\begin{comment}
\begin{mydef}[global latent accuracy specification]
the global latent accuracy of $\mathcal{N}$ is at least $p_0$ if $A_G(\mathcal{N}) \ge p_0.$
\end{mydef}
\end{comment}

In LGA, which requires $D_i$ but not $E_i$, compared to regular accuracy on the holdout set, we have replaced real data samples with generated samples, following class probabilities in $\mathcal{D}^\text{joint}$ (also note that it is possible to consider similar metrics for each class separately).
As a result, an unlimited number of samples can be used to estimate LGA.
In addition, misclassified samples found during the check of this specification can be used to train $\mathcal{N}$ further.
%The maximum value of $p_0$ for which this specification is satisfied can be estimated based on sampling a random class label $i$ and then an item from $\mathcal{D}_L^{i}$ (see Section~\ref{sec:distr_approx}).
In LRA, instead of generating new samples, we take the approximations of real ones computed with both $D_i$ and $E_i$.
%\commentIB{Benefits of this? Distribution shift? Again can use in training.}
This resembles the Defense-GAN~\cite{samangouei2018defense} approach.

%\commentIB{These are quite preliminary metrics... if the paper is positioned not to be exclusively about robustness, then they are OK to be evaluated in the end.
%However, haven't they been introduced in some other works?
%I haven't searched.
%On the other hand, reporting these metrics gives a baseline for more interesting ones.}

While LGA can be measured by sampling latent codes, LRA can be estimated based on samples from the holdout set (see Section~\ref{sec:checking_global_accuracy}).
%Confidence intervals can be computed if needed.
The main purpose of LGA and LRA in this paper is to serve as baselines for other metrics proposed in the following subsections, which, in addition to generation or reconstruction, assume the presence of an adversary.

%Note that this specification, especially in the case of a random class, is similar to the thresholding constraint on the accuracy on the holdout set, except that (1) the data distribution is replaced with an approximation, (2) an unlimited number of samples can be used to estimate $p_0$.

%Unfortunately, the interpretation of this definition has a problem: the fact that $l$ is sampled from $\mathcal{D}_L^{i}$ does not guarantee that $i$ is the \emph{correct} class of $l$ (e.g., according to human judgment).
%This problem is worsened by using an approximated data distribution.

\subsection{Noise corruption robustness in the latent space}
\label{sec:noise_accuracy}

In this subsection, we consider a randomized noise-adding adversary.
Suppose that $N_{\epsilon, l}$ is some \emph{noise-adding distribution} that operates on latent vectors $l$, where parameter $\epsilon \ge 0$ controls the magnitude of the noise.
Below, we will use the same notation for the probability density function (PDF) of this distribution.
We would like the following conditions to be satisfied:
\begin{enumerate}
\item \emph{Distribution preservation}: for all $\epsilon$, sampling $l' \sim N_{\epsilon, l}$ with $l \sim \mathcal{D}_L^{i}$ is equivalent to sampling $l' \sim \mathcal{D}_L^{i}$.
This condition ensures the ``naturality'' of noise: its addition does not shift the distribution of input vectors, meaning that it will not produce vectors that are not plausible according to $\mathcal{D}_L^{i}$ (compared, e.g., with addition of noise to each component of the original data item).
\item \emph{Support of small noise}: if $\epsilon \to 0$, random vectors $X_\epsilon \sim N_{\epsilon, l}$ converge (e.g., in probability) to $X \equiv l$, i.e., the added noise becomes negligible.
This condition ensures that small $\epsilon$ corresponds to small noise.
\item \emph{Support of large noise}: if $\epsilon \to +\infty$, random vectors $X_\epsilon \sim N_{\epsilon, l}$ converge (e.g., in probability) to $X \sim \mathcal{D}_L^{i}$, i.e., the unperturbed latent vector $l$ becomes irrelevant.
This condition ensures that large $\epsilon$ corresponds to large noise.
Convergence to $\mathcal{D}_L^{i}$ is needed to comply with the first condition.
\end{enumerate}

We will propose a concrete family of distributions satisfying these properties in Section~\ref{sec:noise}.
Now, we look at the case where the input to be classified is a perturbed version of the reconstruction of a real data element:

\begin{mydef}[local latent noise accuracy]
the local latent noise accuracy (LLNA) of $\mathcal{N}$ in point $x \in \mathbb{R}^{n_I}$ of known class $i$ with noise magnitude $\epsilon$ is:
$$\mathrm{LLNA}(\mathcal{N}, \epsilon, x, i) \defeq \prob{l \sim N_{\epsilon, E_i(x)}}{\mathcal{N}(D_i(l)) = i}.$$
\end{mydef}

%The probability above is taken over the probabilistic behavior of $N_{\epsilon}^{c=i}$, which has not yet been clarified.

LLNA is similar to LRA, except that checks are performed on noisy reconstructions of a fixed real data sample $x$.
LLNA can be evaluated based on sampling noise vectors (see Section~\ref{sec:noise}).

\subsection{Adversarial robustness in the latent space}
\label{sec:latent_robustness}

Next, instead of checking the classifier's resistance to random noise, we consider perturbations chosen by an adversary.
In terms of $N_{\epsilon, l}$, we can assume that the adversary can choose the worst case input within bounded likelihood.
Given fixed $x$ and $i$, $l' = N_{\epsilon, E_i(x)}$ is a random $n_L$-dimensional vector.
Then:

\begin{mydef}[local latent adversarial robustness]
the local latent adversarial robustness (LLAR) of $\mathcal{N}$ in point $x \in \mathbb{R}^{n_I}$ with known class $i$, with noise magnitude $\epsilon$, is:
$$\mathrm{LLAR}(\mathcal{N}, \epsilon, E_i(x), i) \defeq \max \{\tau \:|\: \forall l' \in \mathbb{R}^{n_L}: (N_{\epsilon, E_i(x)}(l') \ge \tau \Rightarrow \mathcal{N}(D_i(l')) = i)\}.$$
\end{mydef}

%\commentIB{Proof that max exists. With for all x: f(x) >= t, is there always max t for which this is satisfied? Assume that there is no max t, only sup t. We have sequence t_i -> t from below, and t is not the maximum. We have for all x: f(x) >= t_1, >= t_2 ..., but there exists x_0 such that f(x_0) = t_0 < t. At the same time, f(x_0) >= t_i for all i. But then t cannot be the supremum, as for t_0 < t the condition is already not satisfied.}

This defines LLAR as the maximum likelihood $\tau$ of a latent adversarial perturbation and, with low LLAR corresponding to high robustness.
%We speak of a log-likelihood instead of likelihood to simplify work with resulting values of this metric in high-dimensional spaces.
%However, this definition comes with a caveat: LLAR of a point $x$ is defined in terms of likelihood in $N_{\epsilon, E_i(x)}$, which does not generally correspond to the likelihood of $x$ in $\mathcal{D}^{i}$.
%Nonetheless,
LLAR captures proximity in the latent space and is similar to known definitions of local robustness checked in the input space of the ANN~\cite{anderson2019optimization,bastani2016measuring,fawzi2018analysis,huang2017safety,katz2017reluplex,singh2019abstract}, for example, to pointwise robustness~\cite{bastani2016measuring}.
However, the likelihood $\tau$ of a multivariate random vector may be inconvenient to operate with, and thus we allow it to be post-processed with some decreasing function $g_\epsilon(\tau)$.
In Section~\ref{sec:search_pgd}, we will propose an approach that views LLAR as $\ell_2$ robustness in the latent space (i.e., $g_\epsilon$ will convert the likelihood to this norm) and either finds its approximate value or checks whether it is above a given threshold.

Next, we transform LLAR to global performance metrics, returning to the ideas of sampling latent vectors and looping through reconstructed data items:
%We proceed with the following definitions:

\begin{mydef}[latent adversarial generation severity]
the latent adversarial generation severity (LAGS) of $\mathcal{N}$ with noise magnitude $\epsilon$ is:
$$\mathrm{LAGS}(\mathcal{N}, g_\epsilon, \epsilon) \defeq \expectation{i \sim \mathcal{D}^\text{labels},\; l \sim \mathcal{D}_L^{i}}{g_\epsilon(\mathrm{LLAR}(\mathcal{N}, \epsilon, l, i))}.$$
\end{mydef}

\begin{mydef}[latent adversarial reconstruction severity]
the latent adversarial reconstruction severity (LARS) of $\mathcal{N}$ with noise magnitude $\epsilon$ is:
$$\mathrm{LARS}(\mathcal{N}, g_\epsilon, \epsilon) \defeq \expectation{(x, i) \sim \mathcal{D}^\text{joint}}{ {g_\epsilon(\mathrm{LLAR}(\mathcal{N}, \epsilon, E_i(x), i))}}.$$
\end{mydef}

\begin{mydef}[latent adversarial generation accuracy]
the latent adversarial generation accuracy (LAGA) of $\mathcal{N}$ with noise magnitude $\epsilon$ and bound $\rho$ on its transformed likelihood is:
$$\mathrm{LAGA}(\mathcal{N}, g_\epsilon, \rho, \epsilon) \defeq \prob{i \sim \mathcal{D}^\text{labels},\; l \sim \mathcal{D}_L^{i}}{g_\epsilon(\mathrm{LLAR}(\mathcal{N}, \epsilon, l, i)) > \rho}.$$
\end{mydef}

\begin{mydef}[latent adversarial reconstruction accuracy]
the latent adversarial reconstruction accuracy (LARA) of $\mathcal{N}$ with noise magnitude $\epsilon$ and bound $\rho$ on its transformed likelihood is:
$$\mathrm{LARA}(\mathcal{N}, g_\epsilon, \rho, \epsilon) \defeq \prob{(x, i) \sim \mathcal{D}^\text{joint}}{{g_\epsilon(\mathrm{LLAR}(\mathcal{N}, \epsilon, E_i(x), i)) > \rho}}.$$
\end{mydef}

LAGS and LARS are similar to adversarial severity as defined in~\cite{bastani2016measuring}, and LAGA and LARA are similar to adversarial frequency~\cite{bastani2016measuring}.
Intuitively, LAGS and LARS are average LLAR values, while LAGA and LARA measure average success rate of passing a specification of being resistant to sufficiently likely latent perturbations.
In Section~\ref{sec:search_pgd}, we will approximately evaluate all these metrics with sampling and PGD.
The overview of all considered latent space performance metrics is given in Table~\ref{tab:comparison}.

\renewcommand{\arraystretch}{1.3}
\begin{table}[htpb]
\caption{Overview of the proposed latent space performance metrics.}
\begin{center}
\begin{tabular}{lllll}\hline
Metric & Abbreviation & Needs $E_i$ & Adversary & Value range\\\hline
Latent generation accuracy                 & LGA  & No  & No  & $[0, 1]$\\
Latent reconstruction accuracy             & LRA  & Yes & No  & $[0, 1]$ \\
Local latent noise accuracy                & LLNA & Yes & Random noise & $[0, 1]$ \\
Local latent adversarial robustness        & LLAR & Yes & PGD & $\mathbb{R}^+$ \\
Latent adversarial generation accuracy     & LAGA & No  & PGD & $[0, 1]$ \\
Latent adversarial generation severity     & LAGS & No  & PGD & $\mathbb{R}^\text{a}$ \\
Latent adversarial reconstruction accuracy & LARA & Yes & PGD & $[0, 1]$ \\
Latent adversarial reconstruction severity & LARS & Yes & PGD & $\mathbb{R}^\text{a}$ \\
\hline
\multicolumn{5}{l}{${}^\text{a}$ May be more restricted depending on the choice of $g_\epsilon$.}
\end{tabular}
\label{tab:comparison}
\end{center}
\end{table}

\section{Evaluating latent space performance metrics}
\label{sec:measuring_metrics}

This section proposes concrete approaches to calculate the values of the metrics defined in Section~\ref{sec:metrics}.
%In order to also address likelihood robustness later in Section~\ref{sec:part_likel}, the first stages of the approach will be made more specific than needed for probabilistic robustness only.
The general idea is to work with the standard multivariate Gaussian distribution as the latent one due to its well-known properties.
This is especially important for addressing latent adversarial robustness in Section~\ref{sec:search_pgd}.

\subsection{Choice of generative models}
\label{sec:distr_approx}

To be able to work with probability densities in the latent spaces $\mathcal{D}_L^{i}$, we need to fix the selection of these spaces.
We achieve this by taking $\mathcal{D}_L^{i} = N(0, I)$.
Then, to evaluate all metrics proposed in Section~\ref{sec:metrics}, transformations $D_i$ and, for reconstruction-based metrics, $E_i$ must be defined for all classes $1 \le i \le m$.
The following techniques can be applied:
\begin{enumerate}
\item
For each $i$, train a generative autoencoder $(\mathcal{N}_i^E, \mathcal{N}_i^D)$ and take $E_i(x) \defeq \mathcal{N}_i^E(x)$, $D_i(l) \defeq \mathcal{N}_i^D(l)$.
\item
For each $i$, train a GAN with generator $\mathcal{G}_i$ and take $E_i(x) \defeq \mathcal{G}_i(x)$.
$D_i$ can be obtained by enhancing these GANs with encoding procedures, e.g., by training inverters~\cite{hendrycks2019natural}, performing gradient-based optimization of latent codes, or both~\cite{bau2019seeing}.
Instead of training models for each class separately, it is possible to train class-conditional models~\cite{odena2017conditional}.

\end{enumerate}

\subsection{Measuring latent accuracy}
\label{sec:checking_global_accuracy}

With $D_i$ and $E_i$ defined, \textbf{LGA can be measured} by repeatedly sampling a class label $i \sim \mathcal{D}^\text{labels}$ and a latent code $l \sim N(0, I)$, calculating $o_g = [\mathcal{N}(D_i(l)) = i]$,\footnote{$[x]$ (Iverson bracket) is 1 if $x$ is true, and 0 if $x$ is false.} which is a Bernoulli random variable, and averaging the obtained values of $o_g$, which gives an unbiased estimate of LGA.
Similarly, \textbf{LRA can be measured} by sampling validation data items $(x, i)$ and averaging $o_r = [\mathcal{N}(D_i(E_i(x)) = i]$.

%Its expectation is the maximum value of $p_0$ for which the specification is satisfied.
%We can estimate it as sample mean $\hat{p}_0$ and compute a confidence interval based on $\hat{p}_0$ and the number of samples.
%If the width of the confidence interval is predefined, the number of samples can be increased until it is reached.

\subsection{Noise model and measuring local latent noise accuracy}
\label{sec:noise}

Suppose that we sample $(x, i) \sim \mathcal{D}^\text{joint}$ by enumerating over $(x_1, y_1), ..., (x_k, y_k)$.
%, and $i \in \{1, ..., m\}$ is the label of $x \in \mathbb{R}^{n_I}$.
In this case $l = E_i(x) \sim N(0, I)$.
%, where $s$ and $A$ correspond to the distribution $\mathcal{D}^{c = y}$ and $s$ is a component-wise application of $s_j$ ($1 \le j \le n_L$).
At this point, we can inject a random perturbation into the latent code.
We define the noise-adding distribution $N_{\epsilon, l}$ in the following way:
\begin{equation}
l' \sim N_{\epsilon, l} \defequiv l' = \frac{l + \epsilon \cdot \delta l}{\sqrt{1 + \epsilon^2}} \text{ with } \delta l \sim N(0,  I).
\label{eq:noise_original}
\end{equation}
Note that, given the previous choice $\mathcal{D}_L^{i} = N(0, I)$, this choice of $N_{\epsilon, l}$ complies with the constraints stated in Section~\ref{sec:problem}, and it would not be distribution-preserving either (1) with a non-Gaussian $\delta l$, or (2) without the denominator $\sqrt{1 + \epsilon^2}$.
Furthermore, the definition~(\ref{eq:noise_original}) is equivalent to:
\begin{equation}
N_{\epsilon, l} = N\left(\frac{l}{\sqrt{1 + \epsilon^2}}, \frac{\epsilon^2}{1 + \epsilon^2}I\right).
\label{eq:noise}
\end{equation}

\textbf{LLNA can be measured} as follows: find the latent vector $l = E_i(x)$, then repeatedly sample $l' \sim N_{\epsilon, l}$ and calculate $o_n = [\mathcal{N}(l') = i]$, which is again a Bernoulli random variable.
The rest is similar to checking LGA and LRA.

\subsection{Likelihood of perturbations and perturbed vectors}
\label{sec:pert_likelihood}

In the rest of this section, to check LLAR and its derivatives, we will optimize \emph{latent perturbations}---adversarially chosen perturbations that are bounded by the likelihood of the outcomes of $N_{\epsilon, l}$.
They are similar to the ones considered in~\cite{zhao2017generating}.
Noise addition $N_{\epsilon, l}$ (\ref{eq:noise}) can be interpreted as a composition of two transformations:
\begin{enumerate}
\item \emph{decay} (reduction) of the unperturbed latent vector $l$ by $\sqrt{1 + \epsilon^2}$;
\item addition of Gaussian noise $\Delta l \sim N\left(0, \epsilon^2 I / (1 + \epsilon^2)\right)$.
\end{enumerate}

%Consider a simpler noise-adding procedure that does not include scaling by $\sqrt{1 + \epsilon^2}$ as in Section~\ref{sec:noise}:
%$$N_{\epsilon}^{c=y}(l) = l + \Delta l, \text{ where } \Delta l \sim N(0, \epsilon^2I).$$

%\todo{Is scaling by $\sqrt{1 + \epsilon^2}$ meaningful when searching for PPAPs? It can be used, but only adds complexity and nothing essential. The scaled distribution is again Gaussian. And below, even $\epsilon$ becomes irrelevant.}

Below, we will refer to $\Delta l$ as a \emph{latent adversarial perturbation} rather than noise, emphasizing that $\Delta l$ will be produced with directed search rather than sampling.
What perturbations $\Delta l$ are more likely?
The log-likelihood of $\Delta l$ having a standard Gaussian distribution is determined by the $\ell_2$ norm of $\Delta l$:
\begin{multline}
\log f_{N\left(0, \epsilon^2 I / (1 + \epsilon^2) \right)}(\Delta l) = \log \prod_{j = 1}^{n_L} \sqrt{\frac{1 + \epsilon^2}{2 \pi \epsilon^2}} \exp\left(-\frac{1 + \epsilon^2}{2 \epsilon^2}\Delta l_j^2\right)\\
= n_L \log\sqrt{\frac{1 + \epsilon^2}{2 \pi \epsilon^2}}  - \frac{1 + \epsilon^2}{2 \epsilon^2}\sum_{j = 1}^{n_L} \Delta l_j^2 = c_1(\epsilon) - c_2(\epsilon) \norm{\Delta l}_2^2.$$
\label{eq:likelihood}
\end{multline}

The distribution of the perturbed vector $l' = l / \sqrt{1 + \epsilon^2} + \Delta l$, which is of interest in the definition of LLAR, differs from the one of $\Delta l$ only by its mean, and thus its log-likelihood as a function of $\Delta l$ is the same.

%However, the primary goal of finding adversarial perturbations is to change the classification decision of $\mathcal{N}$.
%If this is impossible given the constraints on possible perturbations, then at least we would like to increase the classification loss $L(\mathcal{N}, D_i(l'), i)$, where $i$ is its correct class label, as much as possible.

\subsection{Optimization problem for bounded latent perturbation search}

%Typically, adversarial perturbations are searched with PGD~\cite{madry2017towards}.
%Most commonly, studies consider $\ell_{\infty}$ (the maximum absolute value of the vector's components) and $\ell_2$ norms to constrain adversarial perturbations.
%In~\cite{moosavi2016deepfool}, the DeepFool algorithm for finding a minimum perturbation according to $\ell_p$ (and in particular $\ell_2$) norms is proposed, which is a form of a gradient descent that is intended to converge to a solution near the decision boundary of the classifier.

To measure LAGA and LARA (Section~\ref{sec:latent_robustness}), it is sufficient to check whether LLAR at the current latent point is bounded with a defined likelihood $\tau$ (according to the noise model from Section~\ref{sec:noise}): that is, any perturbation whose likelihood is at least $\tau$, is class-preserving.
According to Eq.~\ref{eq:likelihood}, each positive value $\tau$ uniquely corresponds to a particular value of the $\ell_2$ norm of the perturbation $\Delta l$ around $l / \sqrt{1 + \epsilon^2}$.
For convenience, we will measure perturbation likelihood with its \emph{scaled norm} %\footnote{A norm scaled by a positive constant is also a norm.}
$\scnorm{\cdot} = \norm{\cdot}_2 / \sqrt{n_L}$.
With this scaling, the expected squared scaled norm of a multidimensional vector distributed according to $N(0, I)$ is one.
The following function transforms the likelihood of $\Delta l$ to $\scnorm{\Delta l}$:
$$g_\epsilon(\tau) = \sqrt{\frac{c_1(\epsilon) - \log \tau}{n_L \cdot c_2(\epsilon)}}.$$

We also introduce the following auxiliary definitions:
\begin{itemize}
\item $l_0$ is the initial latent vector, where a LLAR specification should be checked.
It corresponds to some input vector $x$ with its available label $i$: $l_0 = E_i(x)$.
\item The \emph{decay factor} $d = 1 - 1 / \sqrt{1 + \epsilon^2}$ ($0 \le d \le 1$) is the amount of reducing the vector $l$ prior to the search of a perturbation.
\item $l_1 = (1 - d) l_0 = l_0 / \sqrt{1 + \epsilon^2}$ is the reduced vector, which is the mean of the perturbation $\Delta l$.
%\footnote{More precisely, this value is a $\chi$-distributed random variable, whose mean
%can be approximated as $\sqrt{n_L}$ for $n_L \gg 1$.
%is $\mu = \sqrt{2} \Gamma((n_L + 1) / 2) / \Gamma(n_L / 2)$. $\sqrt{n_L}$ is an approximation to $\mu$ for $n_L \gg 1$, which is suitable for our purposes.
%}
\end{itemize}

Thus, we need to check whether there is an adversarial perturbation $\Delta l$ with $\scnorm{\Delta l} \le \rho$, where $\rho = g_\epsilon(\tau)$,
%(note that large $\tau$ correspond to small $\rho$ and vice versa)
that makes the classifier $\mathcal{N}$ classify $D_i(l_1 + \Delta l)$ as not belonging to class $i$.
Suppose that an \emph{objective function} $\mathcal{O}: \mathbb{R}^{n_L} \to \mathbb{R}$ is available such that $\mathcal{O}(\Delta l) > 0$ implies correct classification and $\mathcal{O}(\Delta l) < 0$ implies misclassification.
We take
$$\mathcal{O}(\Delta l) = s(i) - \max_{1 \le j \le m, j \ne i} s(j), \text{~where~} s(j) = S_\mathcal{N}(D_i(l_1 + \Delta l), j).$$
It is almost everywhere differentiable due to the corresponding assumptions on $S_\mathcal{N}$ and $D_i$.
Then we can solve the following constrained optimization problem with gradient-based techniques:
\begin{equation}
\minimize_{\Delta l : \: \scnorm{\Delta l} \le \rho} \mathcal{O}(\Delta l).
\label{eq:opt_problem}
\end{equation}

%When $m = 2$, we can take $\mathcal{O}(\Delta l) = -L(\mathcal{N}, D_i(l_1 + \Delta l), i)$, which is almost everywhere differentiable due to the corresponding assumptions on $L, \mathcal{N}$, and $D_i$.
%However, for $m > 2$ classes, ...
%Our problem does not distinguish classes different from $i$
%Let $\mathcal{L}(\Delta l)$ denote the loss of $\mathcal{N}$ as a function of the perturbation:
%$$\mathcal{L}(\Delta l) = L(\mathcal{N}, D_i(l_1 + \Delta l), i).$$
%We may assume that $\mathcal{L}$ is almost everywhere differentiable due to the corresponding assumptions on $L, \mathcal{N}$, and $D_i$, and thus we can perform first order maximization of this loss.
%Specifically, we need to solve the following constrained optimization problem:

%\commentIB{Actually, I optimize a refined criterion: $\max_{j \ne i} p_j - p_i$. Introduce soft classification scores? Also, we need to state clearly that we are interested in adversarial examples classifier to any class different from the reference one.}

\subsection{Intuition for non-zero decay factor}

%\commentIB{This explanation covers only the case of using the same autoencoder.}

At first glance, viewing latent perturbations as a perturbation of $l_1$ but not $l_0$ (which equals $l_1$ only in the case of zero noise) may be confusing.
The intuitive explanation, on the other hand, is in line with the purpose of division by $\sqrt{1 + \epsilon^2}$ in (\ref{eq:noise_original}), which is needed to reduce the covariance matrix of the distribution of perturbed vectors (with unperturbed vectors $l \sim N(0, I)$) back to $I$.
\begin{comment}
Assume, for simplicity, that the same autoencoder is used for all conditional latent distributions.
In this case, perturbation search around $l_0$ has a drawback: the chosen perturbation may drive the perturbed latent vector $l$ towards a less likely region in distributions $\mathcal{D}_L$ and $\mathcal{D}_L^{c = y}$.
%Such a movement in $\mathcal{D}_L$ in practice can lead to the adversarial example (e.g., image) looking distorted.
%As for $\mathcal{D}_L^{c = y}$, such a movement may also
Such a movement may \red{change the real class of $l$} while retaining $l$ plausible according to $\mathcal{D}_L$.
When the original vector $x$ (not its latent code) is perturbed, many works~\red{[...]} achieved finding adversarial examples that are almost indistinguishable from the original images by a human---hence, we may assume that the actual label of a perturbed image remains unchanged.
With the latent code being perturbed, the perturbations are more distinguishable and \red{may change the class perceived by a human}.
%The proposed extended algorithm aims to counter exactly this problem: its regularization part increases the class-specific features of $x$.
\end{comment}
Decay moves the search region to the area of more likely (having a smaller norm) vectors.
Again, we remind that the likelihood in $\mathcal{D}_L^{i}$ in the general case does not correspond to the likelihood in $\mathcal{D}^{i}$.
Still, in our experiments, decay moves latent vectors towards ``averaged'' representatives of each class.

%When $\kappa > 0$ (equivalently, $0 < d \le 1$), $l_*$ has a higher likelihood than $l_0$.
%Now, it is best to operate in $\mathcal{D}_L^{c = y}$, where the movement closer to zero moves the vector to the ``most typical'' representative of its class---this is intended to counter the effect of perturbation search which may \red{try to change the actual label of the sample}.
%\commentIB{Limitation: if the original distribution is multimodal, then ``most typical'' may not correspond to the modes and can actually have a low PDF value in the original distribution. The same applies to the approximation of the autoencoder: the maxima of the PDF of the latent code distributions may not correspond to the maxima of the original data distribution.}

\subsection{Latent perturbation search with PGD}
\label{sec:search_pgd}

\begin{figure}[ht]
\centering
\includegraphics[width=0.35\textwidth]{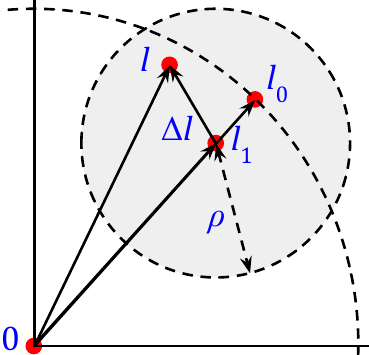}
\caption{Graphical interpretation of latent perturbation search with PGD.
The grey circle is the region where the adversarial perturbation $\Delta l$ is searched, and $l$ is the current candidate solution.}
\label{fig:pgd_search}
\end{figure}

The constrained problem~(\ref{eq:opt_problem}), considered for an approximation $E_y(x)$ of a data element $(x, y)$, corresponds to \textbf{checking a threshold specification on LLAR}.
Our proposed untargeted attack that solves this problem is a variant of PGD~\cite{madry2017towards}.
PGD is started from a random latent perturbation within the allowed $\rho$-ball and is run until a misclassification is achieved, i.e., $\mathcal{O}(\Delta l) < 0$, but no longer than for a predetermined number of steps.
The learning rate is set to ensure that the boundary of the $\rho$-ball can be reached from any point inside it.
To avoid exploding or vanishing gradients, as in~\cite{madry2017towards}, we scale $g = \nabla \mathcal{O}(\Delta l)$ with its $\ell_2$ norm (specifically, we divide $g$ by $\scnorm{g}$).
The optimization procedure is illustrated in Fig.~\ref{fig:pgd_search}.

Next, we consider evaluation of performance metrics that are based on LLAR.
\textbf{Evaluation of LAGA and LARA} is similar to the one of LGA and LRA, except that a generated or approximated point is altered by the PGD adversary.
To increase reliability, PGD should be run multiple times.
To \textbf{evaluate LAGS and LARS}, minimum perturbation bounds $\rho$ need to be calculated and averaged.
%---then, since log-likelihood bound $\tau$ and scaled norm bound $\rho$ are connected with an affine function (Eq.~\ref{eq:likelihood}), so do the expectations (however, for simplicity, in Section~\ref{sec:experiments}, we will present LAGS and LARS values in terms of $\rho$).
To approximately find the minimum norm $\rho$ of a class-changing perturbation without pre-setting it, we apply the following techniques:
\begin{itemize}
\item Set $\rho$ to a large value (we use $\rho = 2.5$) and start PGD with a small learning rate at $\Delta l = 0$.
It will reach some solution, whose norm could be used as an approximation for minimum $\rho$.
\item The solution above might be prone to reaching local optima,
%(in our experiments, this happens on MNIST)
which can be mitigated by several restarts from different points.
In this case, to enforce norm minimization, each new restart is done with $\rho$ set to the scaled norm of the previously found solution, and the learning rate is reduced proportionally to the shrinkage of $\rho$.
\end{itemize}

%\commentIB{Also tried running PGD multiple times with $\rho$ shrinking. It is not successful on CelebA, but solves the problem with local optima on MNIST.}

To evaluate LAGS and LARS, we also explored the use of DeepFool~\cite{moosavi2016deepfool}, which is an algorithm to find minimum $\ell_p$ adversarial perturbations.
Essentially, it is a variant of gradient descent with specifically chosen step magnitudes that are intended for fast convergence to a perturbation lying on the decision boundary of the classifier.
%Therefore, like gradient descent with $\lambda = 0$, it can be used to find an approximate minimum value of $\rho$ for which the local likelihood specification is satisfied.
%Although proposed in the context of the problem of image classification, DeepFool can in principle be applied to any classification problem solved by feed-forward ANNs.
Unfortunately, we observed its frequent divergence on our optimization problem (in~\cite{moosavi2016deepfool}, images were manipulated in the original space).
Gradient clipping resumed convergence, although it often cannot be achieved in just a few steps as in~\cite{moosavi2016deepfool}.
Thus, for the lack of apparent benefits of using DeepFool, in our experiments we apply only PGD.

%\commentIB{Sampling with shrinking. This method was proposed in~\cite{zhao2017generating} and maybe should be implemented to compare our work with them. Advantage of this method: it is black box, i.e., does not require the gradient of the classifier to be available. Maybe we can treat this method as a part of a larger ``framework''. But maybe we can just mention it in order to not overcomplicate things.} \commentIB{Since we assume that we are checking our classifier's specifications, we have the gradient... but we don't have it in NLP tasks.}

\section{Experimental evaluation}
\label{sec:experiments}

\subsection{Implementation and experimental setup}
\label{sec:setup}

The proposed framework of evaluating feed-forward ANN classifier performance with generative models was implemented in Python with \texttt{pytorch}.
The code and models used to obtain the results described in this section are publicly available at \url{https://github.com/igor-buzhinsky/latent-space-nn-evaluation}.
As the case studies, we considered the following image classification problems:
\begin{enumerate}
\item \textbf{MNIST}~\cite{lecun1998mnist} digit classification ($m = 10$ classes).
As generative models, for MNIST, we trained a WGAN~\cite{arjovsky2017wasserstein} with $n_L = 64$ for each class, and implemented $E_i$ with gradient descent (Adam with 4 restarts) over latent codes.
Examples of images reconstructed and generated by these models are given in Fig.~\ref{fig:modeldemo} (top).
\item Gender predictions based on face photos, using the \textbf{CelebA}~\cite{liu2015deep} dataset ($m = 2$ classes: 1 = ``female'', 2 = ``male''; images were center-cropped and resized to $128\times128$ pixels).
For CelebA, we trained PIONEER~\cite{heljakka2018pioneer,heljakka2020towards} generative autoencoders for each dataset and class with $n_L = 511$.
Examples of images produced by the models are given in Fig.~\ref{fig:modeldemo} (middle)---note that the visual quality of reconstructed images is somewhat better compared to generated images.
\item Scene type prediction using the \textbf{LSUN}~\cite{yu2015lsun} dataset ($m = 2$ classes: 1 = ``bedroom'', 2 = ``church outdoor''; images were center-cropped and resized to $128\times128$ pixels).
For LSUN scene types, we also trained PIONEER models with $n_L = 511$.
However, as seen from Fig.~\ref{fig:modeldemo} (bottom), except for bedroom reconstructions, the visual quality of images produced by PIONEER models for LSUN is worse compared to CelebA images.
\end{enumerate}

For each of these classification problems, we trained five deep CNN classifiers (see Appendix~\ref{sec:appendix_training} for details):
\begin{enumerate}
\item $\mathcal{N}_\text{UT}$ (``undertrained''): a classifier trained in a usual way, without data augmentation, but only for one epoch (to intentionally achieve lower accuracy);
\item $\mathcal{N}_\text{NR}$ (``non-robust''): the same as above, but trained only several epochs;
\item $\mathcal{N}_\text{CA}$ (``conventional augmentation''): a classifier trained in a usual way, with conventional data augmentation;
\item $\mathcal{N}_\text{R}$ (``robust''): a classifier trained on images corrupted with visible Gaussian noise~\cite{ford2019adversarial};\footnote{This form of training was used instead of more common robust optimization with PGD to reduce computation time.}
\item $\mathcal{N}_\text{B}$ (``both''): a classifier trained with both conventional data augmentation and noise corruption.
\end{enumerate}

\subsection{Performance evaluation using original space metrics}

The performance metrics of the above deep CNN classifiers in the original space are reported in Table~\ref{tab:classifiers}.
From this table, it is visible that, as expected, training with Gaussian noise achieved not only noise corruption robustness but also adversarial robustness, and the latter two are associated.
In addition, a trade-off is visible between the accuracy of the classifiers on clean images (hereinafter, \emph{clean accuracy}) and adversarial robustness, which is in agreement with previous observations~\cite{tsipras2018robustness}.
%Below, we proceed with evaluating latent space performance metrics of these classifiers.

\renewcommand{\arraystretch}{1.3}
\begin{table}[htpb]
\caption{Performance metrics of considered CNN classifiers measured in the original space.
Accuracy was measured on the validation set of each dataset.
For noise accuracy, we report accuracy on images corrupted with standard Gaussian noise with $\sigma = 0.8$.
Adversarial severity~\cite{bastani2016measuring} is reported for $\ell_2$ and $\ell_\infty$ norms scaled by dividing by $\sqrt{n_I}$ and $n_I$ respectively.
It was estimated on 600 images per classifier.
Adversarial perturbations were searched with PGD: for each image, 15 runs were performed with norm threshold shrinkage as explained at the end of Section~\ref{sec:search_pgd}, except for doing this in the original space.
For each PGD run, we used 50 steps of magnitude $0.05\rho$ from a random point, where $\rho$ is the current norm threshold.
For each value series, the best (largest) value is shown in bold.}
\begin{center}
\begin{tabular}{llrrrrr}\hline
\multirow{2}{*}{Dataset} & \multirow{2}{*}{Classifier} & \multicolumn{2}{c}{Accuracy} & & \multicolumn{2}{c}{Adversarial severity}\\
& & Clean & Noise & & $\norm{\Delta x} = \scnorm{\Delta x}$ & $\norm{\Delta x} = \norm{\Delta x}_\infty$ / $n_I$\\\cline{1-4}\cline{6-7}
\multirow{5}{*}{MNIST}
& $\mathcal{N}_\text{UT}$ &       98.2\%  &       86.7\%  & &       0.0919  &       0.2094\\
& $\mathcal{N}_\text{NR}$ & \best{99.2\%} &       86.1\%  & &       0.0892  &       0.1848\\
& $\mathcal{N}_\text{CA}$ &       98.7\%  &       93.9\%  & &       0.1124  &       0.2754\\
& $\mathcal{N}_\text{R}$  &       99.1\%  & \best{98.5\%} & & \best{0.1702} & \best{0.5018}\\
& $\mathcal{N}_\text{B}$  &       98.3\%  &       97.7\%  & &       0.1687  &       0.4921\\
\cline{1-4}\cline{6-7}
\multirow{5}{*}{CelebA}
& $\mathcal{N}_\text{UT}$ &       95.0\%  &       82.7\%  & &       0.0037  &       0.0094\\
& $\mathcal{N}_\text{NR}$ & \best{97.5\%} &       77.5\%  & &       0.0041  &       0.0099\\
& $\mathcal{N}_\text{CA}$ &       96.5\%  &       48.2\%  & &       0.0033  &       0.0090\\
& $\mathcal{N}_\text{R}$  &       96.8\%  & \best{95.9\%} & & \best{0.0143} & \best{0.0354}\\
& $\mathcal{N}_\text{B}$  &       94.7\%  &       94.0\%  & &       0.0135  &       0.0344\\
\cline{1-4}\cline{6-7}
\multirow{5}{*}{LSUN}
& $\mathcal{N}_\text{UT}$ &       94.3\%  &       74.3\%  & &       0.0025  &       0.0061\\
& $\mathcal{N}_\text{NR}$ & \best{98.0\%} &       50.7\%  & &       0.0031  &       0.0071\\
& $\mathcal{N}_\text{CA}$ &       97.5\%  &       51.5\%  & &       0.0051  &       0.0107\\
& $\mathcal{N}_\text{R}$  &       93.5\%  &       95.8\%  & &       0.0167  &       0.0346\\
& $\mathcal{N}_\text{B}$  &       95.0\%  & \best{96.1\%} & & \best{0.0192} & \best{0.0379}\\
\hline
\end{tabular}
\label{tab:classifiers}
\end{center}
\end{table}

\subsection{Performance evaluation using the proposed latent space metrics}

\begin{figure}[ht]
\centering
\includegraphics[width=\textwidth,trim={0.1cm 6.7cm 0.1cm 0},clip]{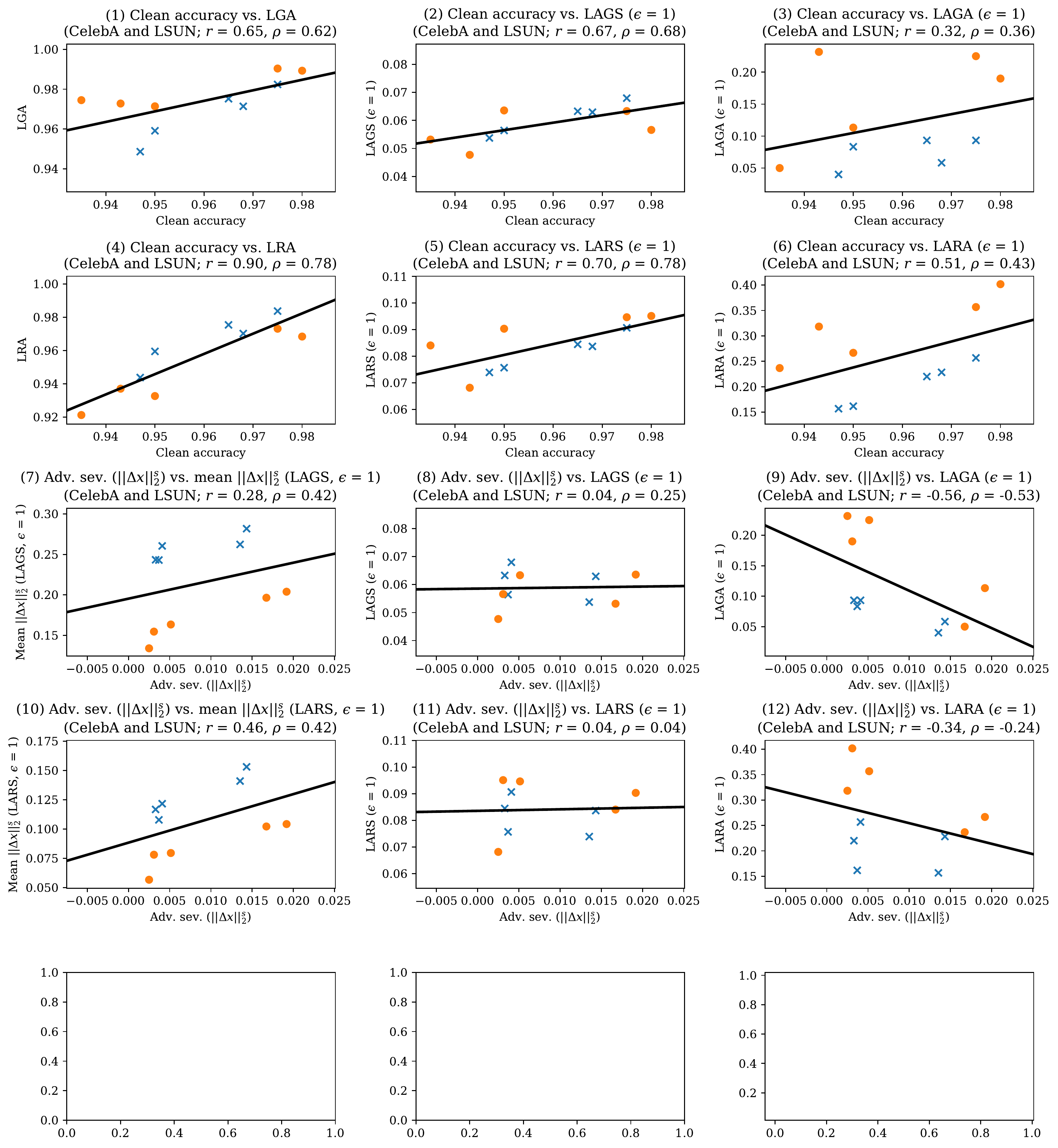}
\caption{Correlation plots for some of the data presented in Table~\ref{tab:robustness_results} (plots 1--6, 8--9, 11--12) and additional data (plots 7 and 10).
Plots are made for CelebA (blue crosses) and LSUN (orange circles) data combined (MNIST data is not shown).
Plots 1 and 4 show the relation between clean accuracy, LRA and LGA.
Plots 2--3 and 5--6 show an association between clean accuracy and latent adversarial robustness (measured as LAGS, LAGA, LARS, LARA).
Plots 7 and 10 show an association between conventional adversarial robustness (measured as adversarial severity with respect to perturbations bounded by scaled $\ell_2$ norm) and the averaged scaled $\ell_2$ norm of found approximately minimum latent perturbations.
Plots 8--9 and 11--12 demonstrate the absence of positive association between conventional adversarial robustness and latent adversarial robustness.
For each plot, Pearson's and Spearman's correlation coefficients ($r$ and $\rho$, respectively) are given.
}
\label{fig:corr_plots}
\end{figure}

We calculated the values of the proposed latent space performance metrics for all aforementioned classifiers.
The corresponding results are provided in Table~\ref{tab:robustness_results} and Fig.~\ref{fig:corr_plots}.
We start interpreting these results from LGA and LRA, which can be regarded as quality measures of generation and reconstruction capabilities of generative models.
For CelebA and LSUN, in Fig.~\ref{fig:corr_plots}, plots~1 and~4, it is visible that clean accuracy is correlated with both LGA and LRA.
The stronger correlation of LRA and clean accuracy can be explained by better reconstruction capabilities of our PIONEER models compared to their generation capabilities.
On MNIST, the associations of clean accuracy with LGA and LRA are roughly the same (Pearson's $r = 0.5$).
Based on these observations, we conclude that the used generative models are suitable for evaluation of other proposed metrics.

\begin{figure}
\begin{tabular}{p{0.94cm}p{0.94cm}p{0.94cm}p{0.94cm}p{0.94cm}p{0.99cm}p{0.94cm}p{0.94cm}p{0.94cm}p{0.94cm}p{0.94cm}p{0.94cm}}
original & $\epsilon = 0$ & $\epsilon = \frac{1}{4}$ & $\epsilon = \frac{1}{2}$ & $\epsilon = \frac{3}{4}$ & $\epsilon = 1$
&
original & $\epsilon = 0$ & $\epsilon = \frac{1}{4}$ & $\epsilon = \frac{1}{2}$ & $\epsilon = \frac{3}{4}$ & $\epsilon = 1$ \\
\end{tabular}
\centering
\includegraphics[width=0.495\textwidth,trim={0.25cm 0.25cm 0.25cm 0.25cm},clip]{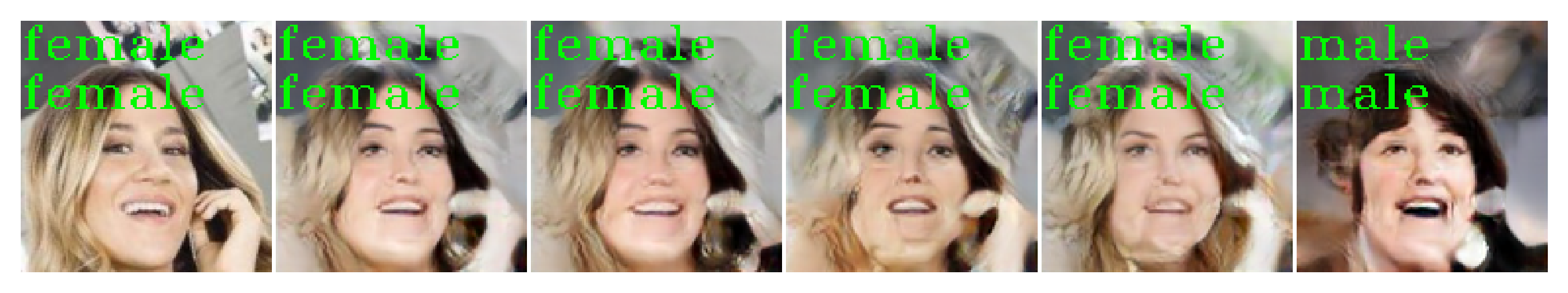} \includegraphics[width=0.495\textwidth,trim={0.25cm 0.25cm 0.25cm 0.25cm},clip]{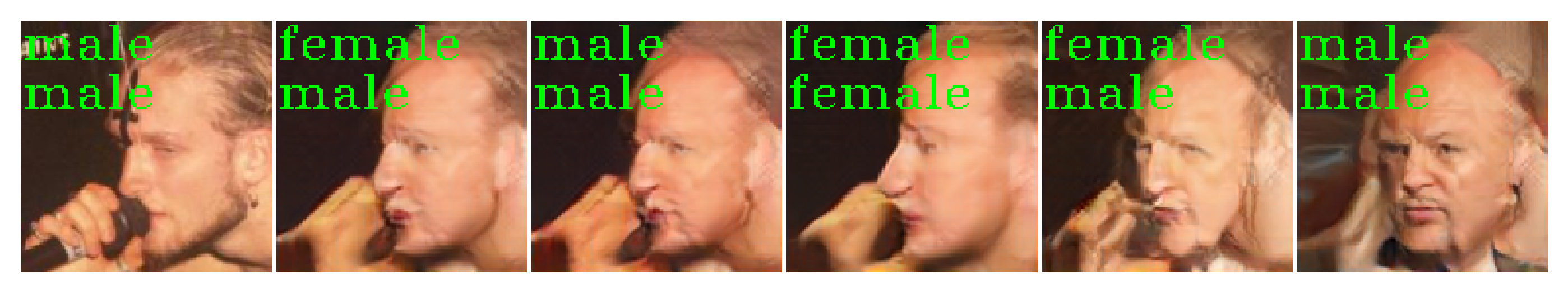}\vspace{-0.1cm}
\includegraphics[width=0.495\textwidth,trim={0.25cm 0.25cm 0.25cm 0.25cm},clip]{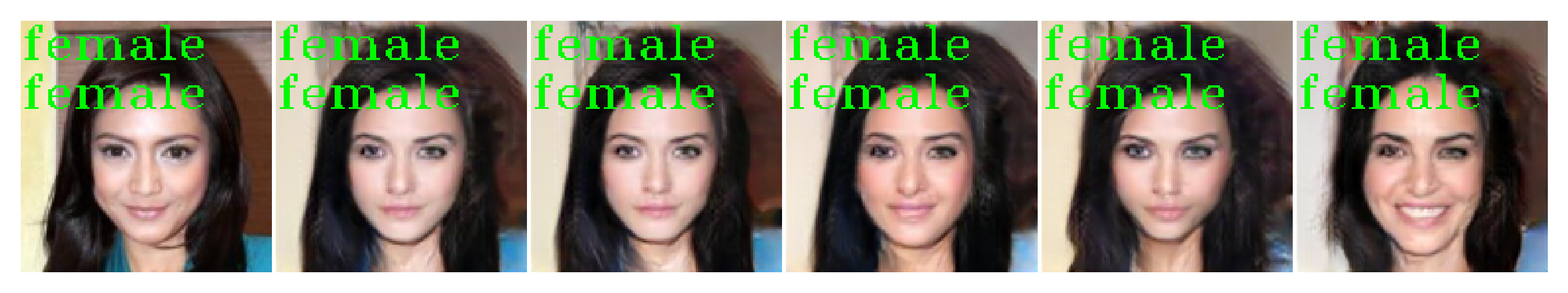} \includegraphics[width=0.495\textwidth,trim={0.25cm 0.25cm 0.25cm 0.25cm},clip]{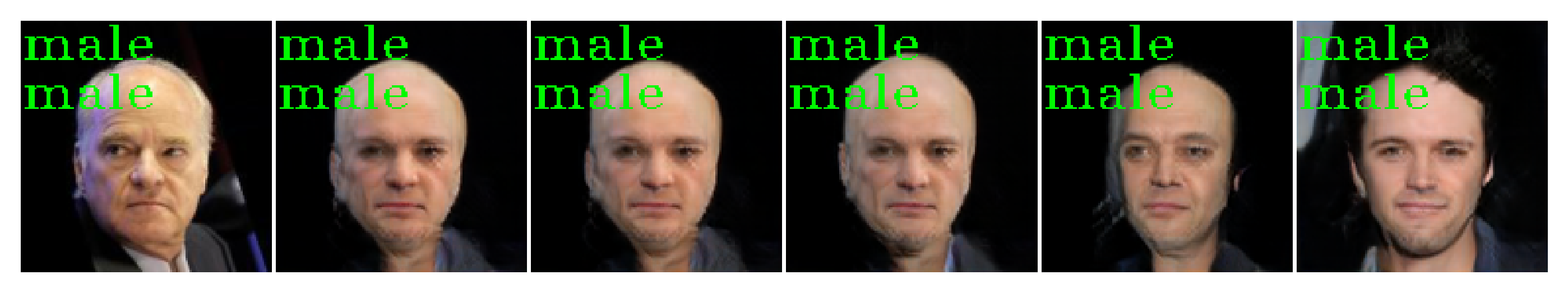}
\includegraphics[width=0.495\textwidth,trim={0.25cm 0.25cm 0.25cm 0.25cm},clip]{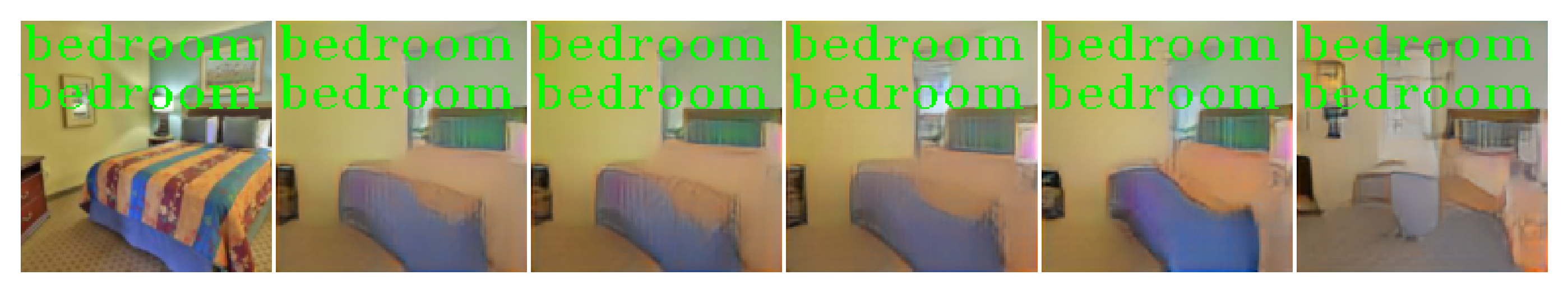} \includegraphics[width=0.495\textwidth,trim={0.25cm 0.25cm 0.25cm 0.25cm},clip]{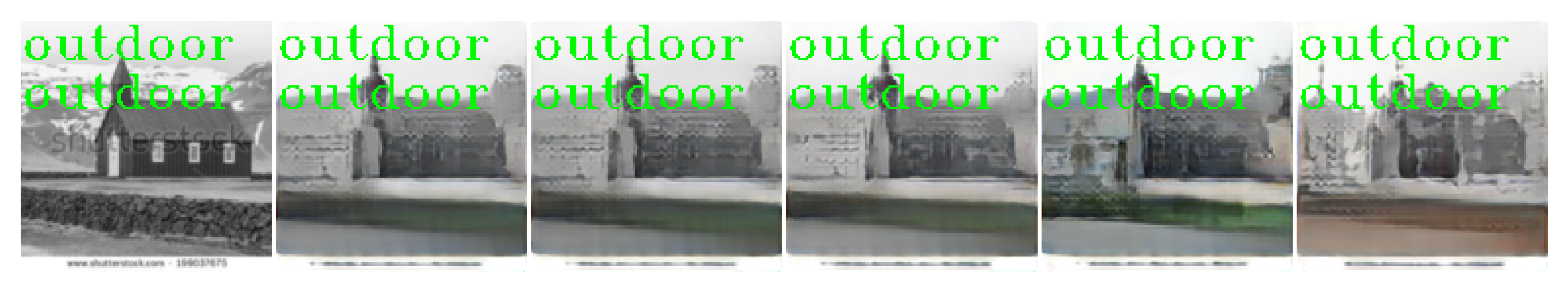}\vspace{-0.1cm}
\includegraphics[width=0.495\textwidth,trim={0.25cm 0.25cm 0.25cm 0.25cm},clip]{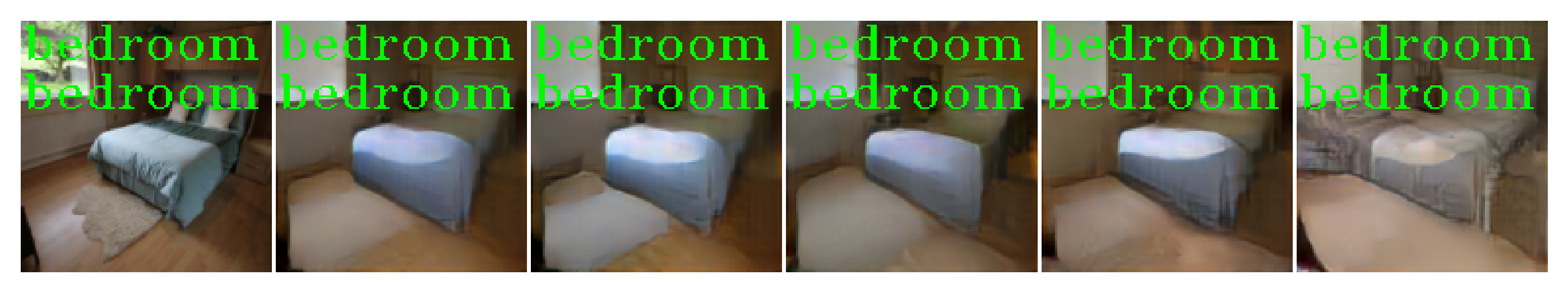} \includegraphics[width=0.495\textwidth,trim={0.25cm 0.25cm 0.25cm 0.25cm},clip]{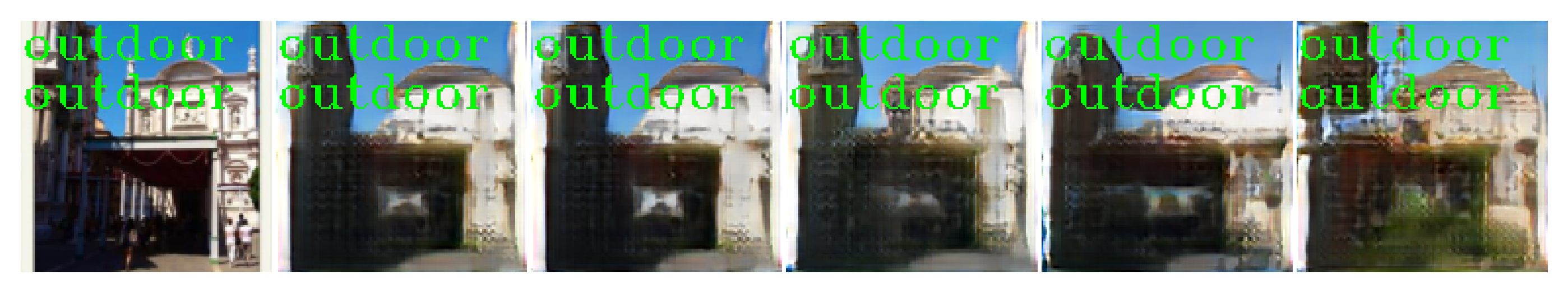}\vspace{-0.1cm}
\caption{Examples of perturbations for CelebA and LSUN images of each class (left: ``female'', ``bedroom'', right: ``male'', ``church outdoor'') that were generated as latent Gaussian noise.
In each image sequence: the original image, the image reconstructed by PIONEER ($\epsilon = 0$), then four perturbed reconstructed images with increasing noise magnitudes $\epsilon = 0.25, 0.5, 0.75, 1.0$.
Green labels show classification outcomes of $\mathcal{N}_\text{NR}$ (on the first line) and $\mathcal{N}_\text{R}$ (on the second line).
All images in this figure have resolution 128$\times$128.}
\label{fig:sampling_perturbations}
\end{figure}

Next, we comment on LLNA, which is a local metric, unlike the others.
We computed its values on particular images and show several noise-based perturbations used in these computations in Fig.~\ref{fig:sampling_perturbations}.
Noise addition appeared to be a very sample-inefficient adversary, but the values of LLNA can be treated as prediction stability measures.
For example, for the reconstructed image in the top right row of Fig.~\ref{fig:sampling_perturbations}, the prediction of $\mathcal{N}_\text{NR}$ is incorrect, and this also reflects in low accuracy of perturbed images (e.g., for $\epsilon = 0.5$, the LNNA on this image is 82.5\%).
The same image is also somewhat difficult for $\mathcal{N}_\text{R}$ (for $\epsilon = 0.5$, LLNA = 92.0\%).
%as well as for its reconstruction, $\mathcal{N}_\text{R}$ predicts the gender incorrectly, but $\text{LLNA}(\mathcal{N}_\text{NR}, 0.25, x, 1) = 65\%$ reveals that this image is challenging also for $\mathcal{N}_\text{NR}$.
%classifier 0
%  initial        : 100.00%
%  reconstructed  : 0.00%
%  perturbed(0.25): 64.50%
%  perturbed(0.50): 82.50%
%  perturbed(0.75): 91.50%
%  perturbed(1.00): 88.00%
%classifier 1
%  initial        : 100.00%
%  reconstructed  : 100.00%
%  perturbed(0.25): 95.00%
%  perturbed(0.50): 92.00%
%  perturbed(0.75): 90.00%
%  perturbed(1.00): 90.00%

The following findings are related to metrics that evaluate adversarial robustness in latent spaces:
\begin{enumerate}
%\item
%The associations of LGA and LRA with clean accuracy can be regarded as quality measures of generation and reconstruction capabilities of generative models, respectively.
\item
We found \textbf{association between clean accuracy and latent adversarial robustness} measured as LAGS, LAGA, LARS, and LARA---see Fig.~\ref{fig:corr_plots}, plots~2--3 and~5--6.
In addition, distribution plots of approximately minimum perturbations found with PGD that were used in computing LAGS and LARS are given in Fig.~\ref{fig:distplots}.
For LARS, examples of such perturbations are shown in Fig.~\ref{fig:latent_pert_example_minimum}, \ref{fig:more_perturbations_celeba} and~\ref{fig:more_perturbations_lsun}.
This finding implies that latent space perturbations may be valuable in training ANN classifiers further.
\item
As visible from Fig.~\ref{fig:latent_pert_example_minimum}, \ref{fig:more_perturbations_celeba}, and \ref{fig:more_perturbations_lsun}, \textbf{latent adversarial perturbations are surprisingly small} on CelebA and LSUN, which indicates that our proposed PGD-based untargeted attack is successful.
At the same time, generated images require smaller latent space perturbations---this can be explained by lower quality of generated images, which makes classifiers less confident in their initial predictions.
On the other hand, on MNIST, perturbations are very large (Fig.~\ref{fig:distplots}, two topmost plots in the first column), significantly raise the norm of the perturbed vector (Fig.~\ref{fig:distplots}, two topmost plots in the second column) and thus exploit the part of the latent space where the generative models were not trained to work.
This can be explained by the simplicity of the MNIST classification problem.
\item
On CelebA and LSUN, we found no association, or even negative association between the conventional adversarial robustness of the classifiers (measured with adversarial severity) and latent adversarial robustness (measured with LAGS, LAGA, LARS and LAGA)---the corresponding plots are given in Fig.~\ref{fig:corr_plots}, plots~8--9 and~11--12.
The outcome of this finding is that latent space performance metrics are different from conventional ones.
This finding is explainable given the association reported in point~1 above and robust classifiers ($\mathcal{N}_\text{R}$ and $\mathcal{N}_\text{B}$) being worse than non-robust ones ($\mathcal{N}_\text{NR}$ and $\mathcal{N}_\text{CA}$) in terms of clean accuracy.
On the other hand, conventional and latent adversarial robustness are correlated on MNIST.
\item
On CelebA and LSUN, the conventional robustness of the classifiers is visible in a different sense: although approximately minimum latent adversarial perturbations for robust classifiers are not larger than the ones for non-robust classifiers in terms of $\scnorm{\cdot}$ in latent spaces, the former correspond to larger image changes measured with $\ell_1$ and $\ell_2$ norms in the original space.
This finding is visible in Fig.~\ref{fig:distplots}, columns~3 and~4.
Larger values of $\ell_1$ and $\ell_2$ norms also imply larger perceptual differences.
%since robust classifiers have slightly lower accuracy, they may need slightly less latent perturbations to be fooled.
%PIONEER models trained for CelebA and LSUN generate images that are visually worse than reconstructed ones.
%Hence, they are more challenging for all classifiers and may need smaller latent perturbations.}
%Actually, to check this, I can change PIONEER settings to favor generation more and train 4 more models, but do I want to do this?
\end{enumerate}

Finally, we confirmed the meaning of decay in the latent space as a countermeasure against the increase of the norm of the latent vector by the adversary: as visible from Fig.~\ref{fig:distplots}, column 2, perturbed vectors typically exceed unperturbed vectors by norm.
This phenomenon is explained by (1) the lower probability density of vectors with large latent space norms and the associated lack of classifier training on such less plausible input images, and (2) a higher ease to exploit a weakness of a generative model with the same sort of vectors.
In particular, the second explanation applies to CelebA, where roughly half of approximately minimum latent space perturbations found with $\epsilon = 0.5~(d = 0.106)$ contained visual artifacts, even though the likelihood of perturbed images in $\mathcal{D}_L^i$ was actually higher than the one of unperturbed images.
With $\epsilon = 1~(d = 0.293)$, the visual quality of perturbed images was higher.
On the other hand, on all datasets, even with $\epsilon = 1~(d = 0.293)$, decayed images were visually close to the originals (this is visible on Fig.~\ref{fig:latent_pert_example_minimum}, \ref{fig:more_perturbations_celeba}, and \ref{fig:more_perturbations_lsun}, columns 2--4).

\renewcommand{\arraystretch}{1.3}
\begin{table}[htpb]
\caption{Latent space performance metrics of considered CNN classifiers.
Accuracy and adversarial robustness computations were performed with 10000 and 600 images respectively.
LARA was measured with $\rho = 0.3$ on MNIST and $\rho = 0.1$ on CelebA and LSUN.
For each value series, the best (largest) value is shown in bold.}
\begin{center}
\begin{tabular}{lrrrrrrrrrrrr}\hline
\multirow{2}{*}{Classifier} & \multicolumn{2}{c}{Accuracy} & & \multicolumn{4}{c}{$\epsilon = 0.5~(d = 0.106)$} & & \multicolumn{4}{c}{$\epsilon = 1.0~(d = 0.293)$}\\
& LGA & LRA & & LAGS & LARS & LAGA & LARA & & LAGS & LARS & LAGA & LARA \\\cline{1-3}\cline{5-8}\cline{10-13}
MNIST\\
$\mathcal{N}_\text{UT}$&      97.7\% &      98.8\% & &      0.3026 &      0.3061 &      37.7\% &      33.8\% & &      0.3117 &      0.3465 &      38.3\% &      41.3\% \\
$\mathcal{N}_\text{NR}$&      98.4\% &\best{99.1\%}& &      0.3264 &      0.3300 &      42.7\% &      41.7\% & &      0.3394 &      0.3613 &\best{50.2\%}&      49.3\% \\
$\mathcal{N}_\text{CA}$&\best{98.5\%}&      98.6\% & &      0.3414 &      0.3373 &\best{46.2\%}&\best{43.2\%}& &      0.3505 &\best{0.3728}&      44.8\% &\best{50.3\%}\\
$\mathcal{N}_\text{R}$ &      98.1\% &      98.9\% & &      0.3385 &      0.3316 &      44.0\% &      40.8\% & &\best{0.3533}&      0.3628 &      47.2\% &      45.5\% \\
$\mathcal{N}_\text{B}$ &      98.2\% &      98.9\% & &\best{0.3510}&\best{0.3385}&      42.3\% &      41.3\% & &      0.3404 &      0.3666 &      42.0\% &      45.3\% \\
\cline{1-3}\cline{5-8}\cline{10-13}
CelebA\\
$\mathcal{N}_\text{UT}$&      95.9\% &      96.0\% & &      0.0582 &      0.0801 &      11.0\% &      22.7\% & &      0.0564 &      0.0757 &       8.3\% &      16.2\% \\
$\mathcal{N}_\text{NR}$&\best{98.2\%}&\best{98.4\%}& &\best{0.0685}&\best{0.0972}&      14.7\% &\best{32.8\%}& &\best{0.0680}&\best{0.0907}& \best{9.3\%}&\best{25.7\%}\\
$\mathcal{N}_\text{CA}$&      97.5\% &      97.5\% & &      0.0661 &      0.0914 &\best{16.0\%}&      30.5\% & &      0.0633 &      0.0845 & \best{9.3\%}&      22.0\% \\
$\mathcal{N}_\text{R}$ &      97.1\% &      97.0\% & &      0.0642 &      0.0957 &      10.7\% &      31.5\% & &      0.0629 &      0.0837 &       5.8\% &      22.8\% \\
$\mathcal{N}_\text{B}$ &      94.9\% &      94.4\% & &      0.0555 &      0.0828 &       7.7\% &      24.3\% & &      0.0537 &      0.0739 &       4.0\% &      15.7\% \\
\cline{1-3}\cline{5-8}\cline{10-13}
LSUN\\
$\mathcal{N}_\text{UT}$&      97.3\% &      93.7\% & &      0.0493 &      0.0794 &      21.2\% &      36.3\% & &      0.0477 &      0.0682 &\best{23.2\%}&      31.8\% \\
$\mathcal{N}_\text{NR}$&      98.9\% &      96.8\% & &      0.0619 &\best{0.1062}&\best{24.3\%}&\best{47.3\%}& &      0.0566 &\best{0.0952}&      19.0\% &\best{40.2\%}\\
$\mathcal{N}_\text{CA}$&\best{99.0\%}&\best{97.3\%}& &\best{0.0665}&      0.1037 &      23.5\% &      43.5\% & &      0.0633 &      0.0947 &      22.5\% &      35.7\% \\
$\mathcal{N}_\text{R}$ &      97.5\% &      92.1\% & &      0.0588 &      0.0976 &       5.2\% &      36.7\% & &      0.0532 &      0.0841 &       5.0\% &      23.7\% \\
$\mathcal{N}_\text{B}$ &      97.1\% &      93.3\% & &      0.0637 &      0.1017 &      10.5\% &      40.2\% & &\best{0.0636}&      0.0904 &      11.3\% &      26.7\% \\
\hline
\end{tabular}
\label{tab:robustness_results}
\end{center}
\end{table}

\newcommand{\pertheader}[0]{
\begin{tabular}{|p{1.62cm}|p{1.62cm}|p{1.62cm}|p{1.62cm}|p{1.62cm}|p{1.62cm}|p{1.62cm}|p{1.62cm}|}
\multicolumn{2}{|c}{Reconstruction} & \multicolumn{2}{|c}{Decay} & \multicolumn{2}{|c}{Perturbation for $\mathcal{N}_\text{NR}$} & \multicolumn{2}{|c|}{Perturbation for $\mathcal{N}_\text{R}$} \\
$x$ & $x_0$ & $x_1$ & $\Delta x$ & $x'_\text{NR}$ & $\Delta x^{}_\text{NR}$ & $x'_\text{R}$ & $\Delta x^{}_\text{R}$ \\
\end{tabular}
}

\begin{figure}
\pertheader
\centering
\includegraphics[width=1.0\textwidth,trim={0.25cm 0.25cm 0.25cm 0.25cm},clip]{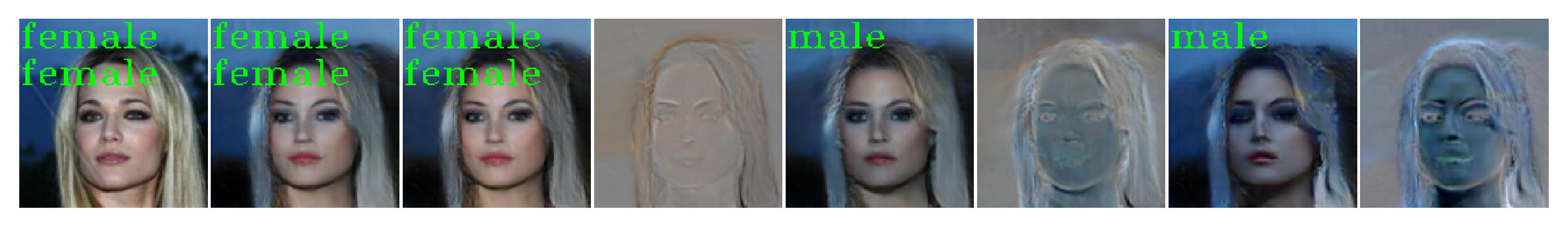}\vspace{-0.1cm}
\includegraphics[width=1.0\textwidth,trim={0.25cm 0.25cm 0.25cm 0.25cm},clip]{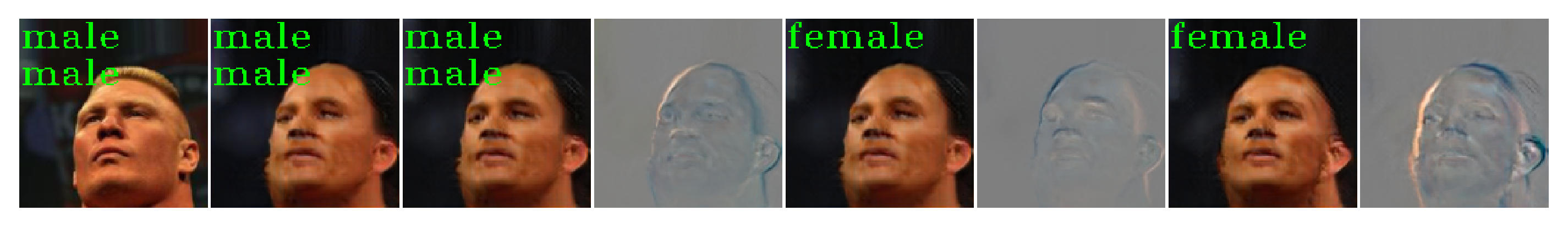}\vspace{-0.1cm}
\includegraphics[width=1.0\textwidth,trim={0.25cm 0.25cm 0.25cm 0.25cm},clip]{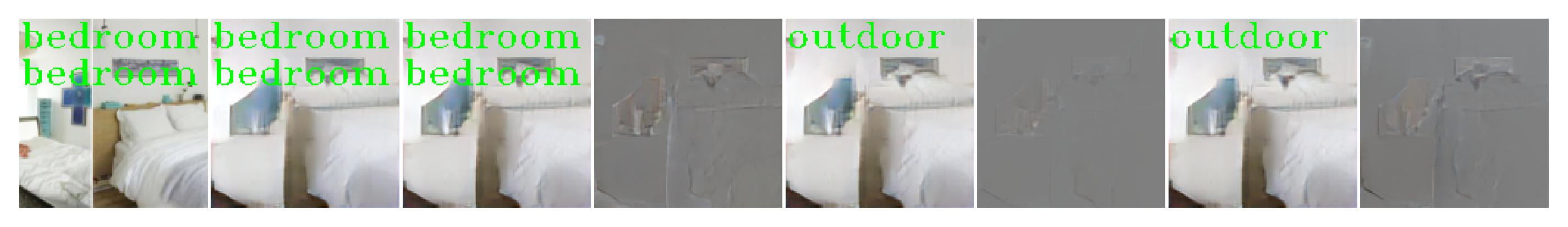}\vspace{-0.1cm}
\includegraphics[width=1.0\textwidth,trim={0.25cm 0.25cm 0.25cm 0.25cm},clip]{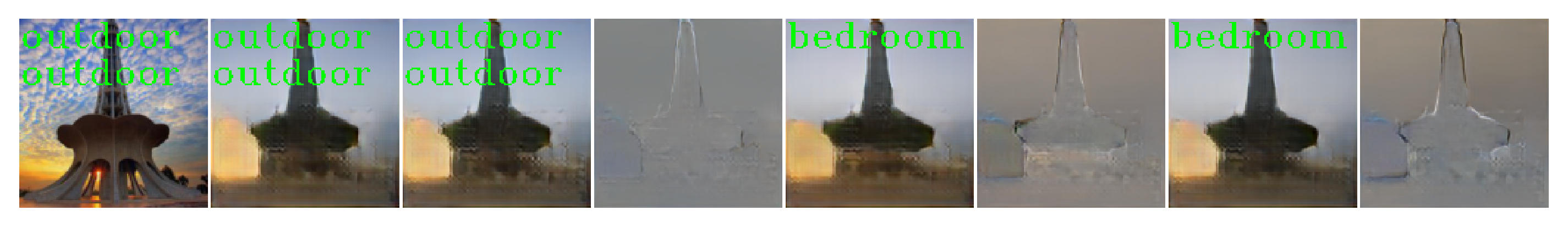}
\caption{Examples of approximately minimum latent CelebA and LSUN image perturbations with $\epsilon = 1~(d = 0.293)$, each found with a single run of PGD from $\Delta l = 0$, for classifiers $\mathcal{N}_\text{NR}$ and $\mathcal{N}_\text{R}$.
In each row, images are given in the following order:
$x$, the real image (with classification outcomes of $\mathcal{N}_\text{NR}$ and $\mathcal{N}_\text{R}$ shown in green);
$x_0 = D_i(l_0)$, the reconstructed image (with both classification outcomes);
$x_1 = D_i(l_1)$, the decayed image (with both classification outcomes);
$\Delta x = x_1 - x_0$, the difference between two previous images;
$x'_\text{NR} = D_i(l'_\text{NR})$, the perturbed image for $\mathcal{N}_\text{NR}$ (with the classification outcome of $\mathcal{N}_\text{NR}$);
$\Delta x^{}_\text{NR} = x'_\text{NR} - x_1$, the perturbation for $\mathcal{N}_\text{NR}$;
$x'_\text{R} = D_i(l'_\text{R})$, the perturbed image for $\mathcal{N}_\text{R}$ (with the classification outcome of $\mathcal{N}_\text{R}$);
$\Delta x^{}_\text{R} = x'_\text{R} - x_1$, the perturbation for $\mathcal{N}_\text{R}$.
All images in this figure have resolution 128$\times$128.}
\label{fig:latent_pert_example_minimum}
\end{figure}

\subsection{Threats to validity}

Below, we list the identified threats to the validity of our study and comment on them:
%\commentAN{Maybe a subsection will indeed be more logical}

%We identified the following threats to the validity of our study:
%\commentAN{In the following paragraphs (subsection?) we will discuss several possible threats to the validity of our study and address them.}

\begin{enumerate}
\item We worked with CNN classifiers of small size (circa 300 thousand trainable parameters), recognizing a small number of classes and with a rather traditional architecture---it may appear that state-of-the-art classifiers, such as the ones for ImageNet, have different patterns of latent space performance metric values.
Yet, we have checked that (1) for the classifiers that we have studied, a connection between adversarial robustness and noise corruption robustness~\cite{ford2019adversarial} exists, (2) on MNIST and CelebA, our robust classifiers have limited capabilities of image generation~\cite{santurkar2019image}.
On LSUN, we have seen that optimizing class activation of robust classifiers adds qualitatively different features to the image compared to non-robust classifiers, but we have not recognized the resulting images as bedrooms nor outdoors.
\item As we measure latent space adversarial robustness (LAGS, LARS, LAGA, LARA) with imprecise attack approaches, we overestimate the values of these metrics.
This bias might have resulted in our classifiers ranked wrongly according to the computed values.
PGD was shown to work well in the original space~\cite{madry2017towards}, but there is so far no similar set of experiments that confirm this property in latent spaces.
We used PGD with 12 restarts to compensate for the possibility of such a bias.
In certain cases (search of minimum adversarial perturbations on CelebA and LSUN), we used a single PGD run with a smaller learning rate, but in these cases, we had ensured that such runs differ insignificantly from the ones of PGD with restarts in terms of the resulting metric values.
\item On LSUN, the small size of the validation set (600 images) may have resulted in prematurely early stopping of training and imprecise accuracy estimates.
In addition, the corresponding generative models produced random images with visible flaws.
Yet, our observations for this dataset are not very different from the ones for CelebA, and perfect generative models might be hard to achieve on custom datasets.
\item PIONEER (CelebA and LSUN) models are designed to be trained to generate images from normalized latent vectors, and the latent distribution is actually the uniform distribution on the unit sphere instead of the Gaussian.
In this paper, this has led to all reconstructed and generated images having unit scaled norm.
Nonetheless, the decoder was capable of accepting unnormalized latent vectors, and decay still worked intuitively, i.e., by softening prominent features of images.
This effect might have been caused by $N(0, I)$ and the uniform distribution on the unit sphere being very similar in multidimensional spaces: $\ell_2$ norms of high-dimensional standard Gaussians are concentrated around $\sqrt{n_L}$.
\end{enumerate}

\section{Related work}
\label{sec:related}

%\subsection{Specifications for ANNs and their verification}
%\todo{A careful review is needed. \commentST{same comment as above applies here. Your goal is not a survey. a decent related work is enough.}}
%Many works on adversarial perturbations consider \emph{local and global robustness}.
%Works that verify things like that: \cite{huang2017safety,katz2017reluplex,singh2019abstract,anderson2019optimization}.
%The work~\cite{anderson2019optimization} proposed a sound and complete approach.
%Output range analysis and safety of output~\cite{pulina2010abstraction,dutta2017output,ruan2018reachability}.

%\subsection{Search of adversarial examples with $\ell_2$ norm}
%DeepFool~\cite{moosavi2016deepfool} constrains perturbations with $\ell_p$, and in particular with $\ell_2$ norm.
%The approach~\cite{zhao2017generating} searches for latent perturbations with $\ell_p$ norm ($\ell_2$ in experiments).
%In~\cite{madry2017towards}, $\ell_2$ norm was considered among others, and the same gradient normalization was used.
%In~\cite{engstrom2019learning}, $\ell_2$ norm and PGD are used.

\subsection{Adversarial examples in latent spaces}

A number of works used generative models to create adversarial attacks and/or defenses.
The work~\cite{zhao2017generating} proposed an approach to search adversarial examples in the latent space of a GAN, also measuring them with $\ell_2$ norms.
This approach is white-box and is based on directed sampling rather than gradient descent, which makes it applicable to discrete input data, such as in natural language processing tasks.
By contrast, our techniques operate in a black-box setting and only with feed-forward ANNs accepting continuous data.
Nonetheless, (1) being based on gradient descent, our latent perturbation search approach is able to find perceptually smaller perturbations (compared to the ones presented in~\cite{zhao2017generating}), (2) we consider a more general framework of transforming data to the latent space and back, (3) we connect latent adversarial robustness to a ``natural'' model of noise in the latent space and this way motivate the use of the $\ell_2$ norm, (4) we search adversarial examples for larger classification tasks ($128 \times 128$ images compared to $64 \times 64$ in~\cite{zhao2017generating}) and latent spaces (511 dimensions compared to 128 in~\cite{zhao2017generating}), and (5) we focus on computing performance metrics for classifiers and not on finding adversarial examples per se.

In~\cite{song2018constructing}, latent space adversarial examples were created from scratch.
This was done using a class-conditional AC-GAN, and evaluation was in particular done on the CelebA (gender classification) and MNIST datasets.
We also search for adversarial examples based on generated data items, however, (1) again, we do it for images larger than $64 \times 64$, (2) we consider the untargeted attack scenario and use a different approach to generate adversarial examples, (3) our approach is not restricted to AC-GANs, and (4) we focus on computing performance metrics rather than finding adversarial examples.

%Generative adversarial examples~\cite{song2018constructing}: generation of new images, not perturbations, by performing a search in the latent space of AC-GAN (conditional GAN). Evaluation on CelebA and genders. The success of examples is shown by using humans to label them (Amazon MTurk).

Generative models were used as defenses against adversarial attacks~\cite{samangouei2018defense,song2018pixeldefend}.
For example, the Defense-GAN~\cite{samangouei2018defense} approach protects image classifiers from adversarial attacks by replacing their input with an approximation in the latent space of a GAN (similarly to what is done when computing LRA in our work).
This defense was broken in~\cite{ilyas2019robust} with an optimization procedure in the latent space subject to a norm constraint in the original space.
Our results are in line with this work, since we similarly approximate input images using a latent space of a generative model, and are able to find perceptually small perturbations that change the prediction of the classifier.
The same work~\cite{ilyas2019robust} also proposed a defense approach based on the search of pairs of examples that are close in the latent space but are scored completely differently by the classifier, and subsequent augmentation of robust optimization with training on these pairs.

\subsection{Robustness metrics for ANNs and their evaluation}

Usually robustness of ANNs to adversarial attacks is measured relatively to a specific attack success. The work \cite{yu2019interpreting} proposes an improvement over the default accuracy-based approach. By analyzing the decision surfaces of models, the authors note that robust models have smooth decision boundaries. The proposed metric reflects this by rewarding models with smooth decision surfaces.
% $$\psi(x) = \frac{1}{\max_{\delta\in B_\epsilon(x)} D_{KL}(P(x), P(x+\delta))} $$
% where $P(x)$ is a reparameterized version of the original network.

The first paper to formalize the notion of adversarial robustness was~\cite{bastani2016measuring}, where the authors propose several metrics quantifying the network robustness, namely, pointwise robustness, adversarial frequency and adversarial severity (see Section~\ref{sec:usual_perf_metrics}). The authors compute the latter two through pointwise robustness, which is measured by approximation.

Exact pointwise robustness calculations are performed in~\cite{boopathy2019cnn}, although the authors refer to the measure as to the ``lower bound on the image distortion.'' Also, the notion of pointwise robustness is explored in~\cite{fawzi2018analysis}, where the authors derive theoretical upper bounds for it. Authors in~\cite{weng2018evaluating} propose an effective proxy measure of network robustness based on measuring Lipschitz constants, although it has received some criticism~\cite{goodfellow2018gradient}. An alternative method to quantifying global robustness properties of networks is proposed by the authors of~\cite{gopinath2017deepsafe}. The authors develop a clustering algorithm that outputs a set of verified regions---a collection of hyperspheres where the network is guaranteed to produce the same label.

\section{Discussion and conclusions}
\label{sec:conclusion}

In this paper, we presented a framework to evaluate the performance of feed-forward ANN classifiers with the help of generative models.
Within the framework, we proposed several performance metrics, the most interesting of which are related to measuring the robustness of classifiers to perturbations in latent spaces of these generative models.
In addition, we presented techniques to evaluate these metrics for classifiers, including a novel PGD-based untargeted attack.
The main motivation of our work is the property of generative models of mimicking the data distribution.
This property implies that the adversarial perturbations that we consider result in natural data changes.

The proposed metrics allowed us to make several interesting observations regarding deep ANN classifiers.
We computed the values of these metrics on several CNN image classifiers and found an association between the accuracy of the classifiers on clean images and adversarial robustness in latent spaces.
This implies that latent adversarial examples might be useful for further classifier training.
We also found that conventional adversarial robustness does not have a strong impact on its latent counterparts, but it is still reflected in the norms of latent adversarial perturbations in the original space.

A speculative explanation of the found connection between the accuracy and latent adversarial robustness is that the latter measures the vulnerability of the classifier to natural adversarial examples, while the accuracy measures the same for random natural examples.
A similar interdependence of accuracy and robustness to natural adversarial examples of a different kind was experimentally found in~\cite{gu2019using}.
An alternative explanation is based on the work~\cite{ford2019adversarial}, where a connection was shown between conventional adversarial robustness and robustness to corruption with Gaussian noise.
When we move to latent spaces, the former becomes LARS/LARA, and the latter becomes the averaged version of LLNA, which, due to our noise model, is just LRA.
In turn, for a generative model with good reconstruction quality, LRA is highly associated with accuracy.
The finding of the work~\cite{ford2019adversarial} is exact for linear models and was shown to hold on CIFAR-10 and ImageNet nonlinear classifiers.
In our case, we can imagine that the classifier accepts latent representations of class $i$, and is actually a composition of $D_i$ and the original classifier $\mathcal{N}$.
Unfortunately, the same properties were not confirmed to hold for classifiers of this kind, and hence this explanation is speculative as well.

%... \red{Metrics and methods to evaluate them? Framework to assess/analyze the performance of ANN classifier with generative models / latent spaces?}

The majority of the proposed metrics relies on the choices of latent distributions as Gaussians and the corresponding Gaussian noise model for this distribution (Eq.~\ref{eq:noise_original}) that together (1) make the noise preserve the distributions of unperturbed vectors $\mathcal{D}_L^i$ and (2) result in simple likelihood bounds as $\ell_2$ norms.
The choice of Gaussians is very conventional, and there is at least one different possible choice: consider the uniform distribution on the unit sphere and the noise model that adds a random Gaussian vector and then normalizes the resulting vector to unit norm.
This solution would still result in $\ell_2$ vector distances monotonically corresponding to noise likelihood.
Some other choices would not achieve both properties (1) and (2).
For example, a Gamma-distributed latent vector would sum with a Gamma-distributed noise vector and still remain Gamma-distributed, but the likelihood of such vectors is more difficult.
Conversely, taking Laplace distributions would result in $\ell_1$ norm likelihood bounds, but summing the unperturbed vector and the noise would not preserve the distribution family.

%According to our preliminary experiments, it is important that generative models are trained for each class $1 \le i \le m$ independently: otherwise, when all classes are present in the latent space, adversarial examples tend to gain the features of other classes (i.e., become rather image manipulations).

%\commentIB{Also discuss why some perturbations have distortions?
%Possibly in connection with the quality of used generative models.}

In the future, we plan to work in the following directions:
\begin{itemize}
\item Check experimentally whether the findings of the work~\cite{ford2019adversarial} are also satisfied in latent spaces---this would further clarify the relationship between the accuracy and latent adversarial robustness.
\item Perform robust manifold defense~\cite{ilyas2019robust} or other form of training with latent adversarial examples, and explore the impact of this training on the values of performance metrics.
\item The proposed metrics to measure latent space adversarial robustness (LAGS, LARS, LAGA, LARA) can be treated as specifications for ANN classifiers, given a threshold on their values to be satisfied.
Gradient-based approaches of checking them are imprecise, and verification of even simpler ANN properties was proven to be NP-hard~\cite{katz2017reluplex}.
A precise, but more computationally intensive way of checking ANN specifications is formal verification~\cite{anderson2019optimization,dutta2017output,huang2017safety,katz2017reluplex,katz2019marabou,ruan2018reachability,singh2019abstract}.%, and this way of checking specifications may be considered in future work.
%Unfortunately, this will not eliminate the imprecision that comes from the use of imperfect generative models.
\end{itemize}

%\begin{acknowledgements}
\section*{Acknowledgments}

The work was financially supported by the Government of the Russian Federation (Grant 08--08).
We acknowledge the computational resources provided by the Aalto Science-IT project.
We thank Ari Heljakka for his help related to the use of the PIONEER generative autoencoder.

%\end{acknowledgements}

\bibliographystyle{spmpsci}      % mathematics and physical sciences
\bibliography{paper}   % name your BibTeX data base

\newpage
\appendix
\renewcommand\thefigure{\thesection.\arabic{figure}}
\setcounter{figure}{0}

\section{Appendix: additional figures}
\label{sec:appendix_figures}

\begin{figure}[h!]
\centering
MNIST, reconstructed (one image per class; each original image is followed by its reconstruction):\vspace{0.05cm}
\includegraphics[width=1.0\textwidth,trim={0.25cm 0.25cm 0.25cm 0.25cm},clip]{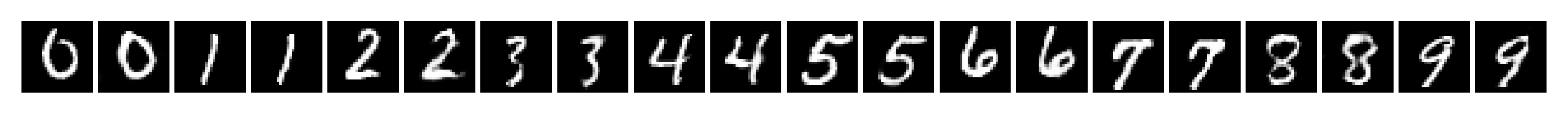}\vspace{0.1cm}
MNIST, generated (two images per class):\vspace{0.05cm}
\includegraphics[width=1.0\textwidth,trim={0.25cm 0.25cm 0.25cm 0.25cm},clip]{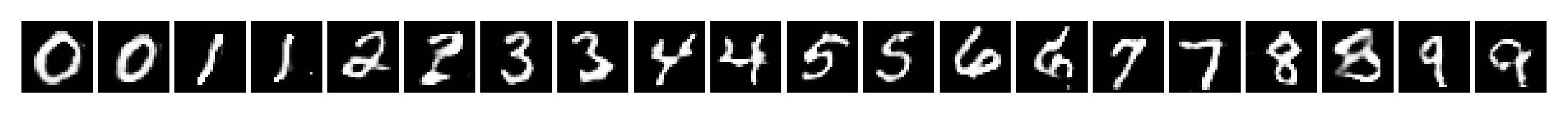}
\noindent\rule{\textwidth}{0.4pt}\vspace{0.06cm}
CelebA (classes ``female'' and ``male''), reconstructed:\vspace{0.05cm}
\includegraphics[width=1.0\textwidth,trim={0.25cm 0.25cm 0.25cm 0.25cm},clip]{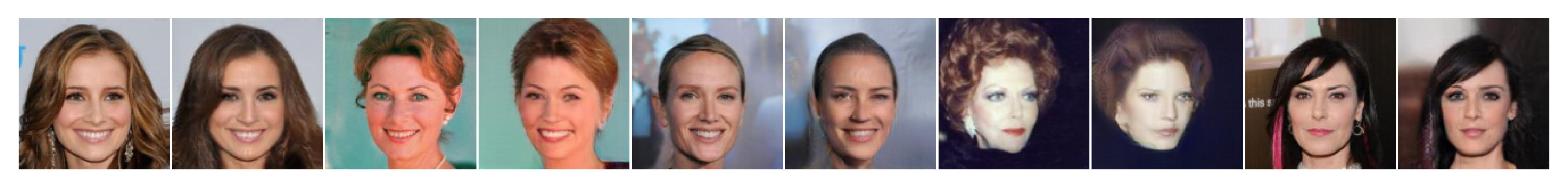}
\includegraphics[width=1.0\textwidth,trim={0.25cm 0.25cm 0.25cm 0.25cm},clip]{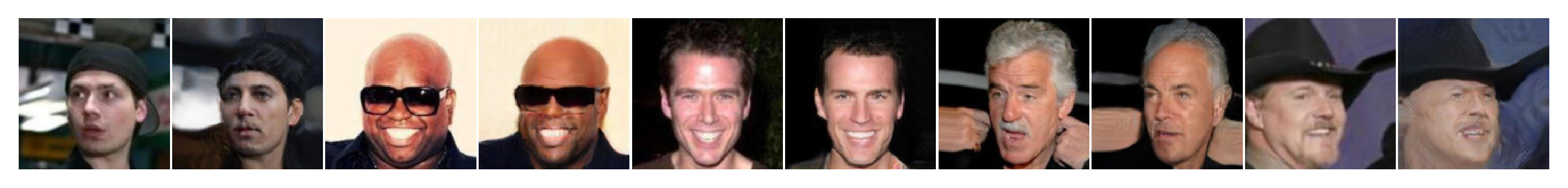}\vspace{0.1cm}
CelebA (classes ``female'' and ``male''), generated:\vspace{0.05cm}
\includegraphics[width=1.0\textwidth,trim={0.25cm 0.25cm 0.25cm 0.25cm},clip]{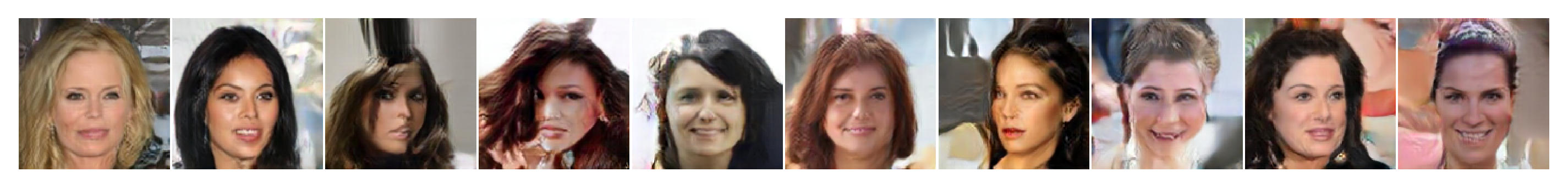}
\includegraphics[width=1.0\textwidth,trim={0.25cm 0.25cm 0.25cm 0.25cm},clip]{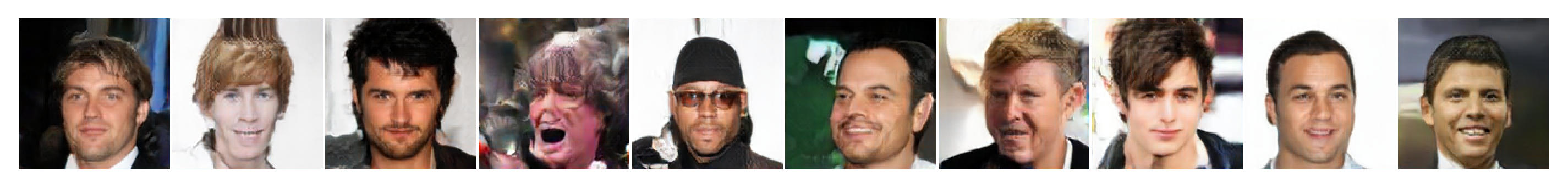}
\noindent\rule{\textwidth}{0.4pt}\vspace{0.06cm}
LSUN (classes ``bedroom'' and ``church outdoor''), reconstructed:\vspace{0.05cm}
\includegraphics[width=1.0\textwidth,trim={0.25cm 0.25cm 0.25cm 0.25cm},clip]{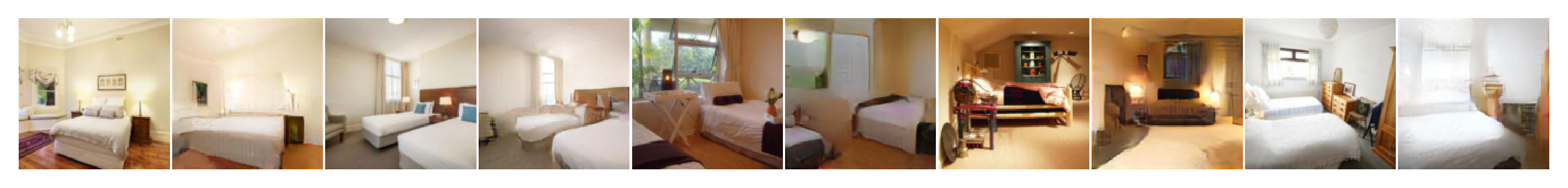}
\includegraphics[width=1.0\textwidth,trim={0.25cm 0.25cm 0.25cm 0.25cm},clip]{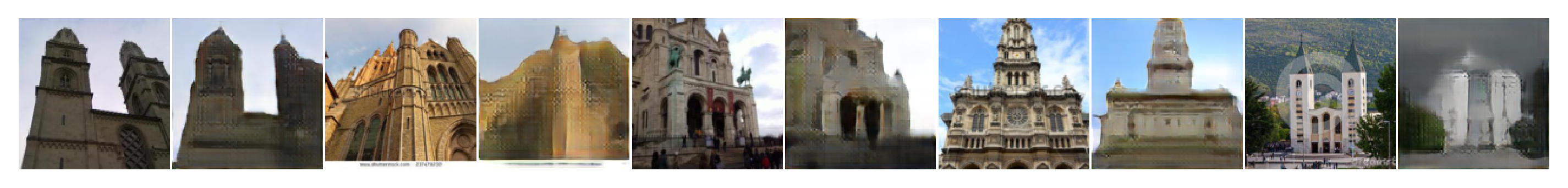}\vspace{0.1cm}
LSUN (classes ``bedroom'' and ``church outdoor''), generated:\vspace{0.05cm}
\includegraphics[width=1.0\textwidth,trim={0.25cm 0.25cm 0.25cm 0.25cm},clip]{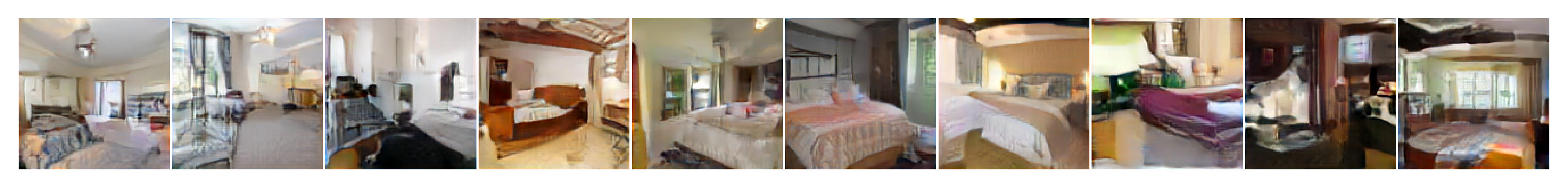}
\includegraphics[width=1.0\textwidth,trim={0.25cm 0.25cm 0.25cm 0.25cm},clip]{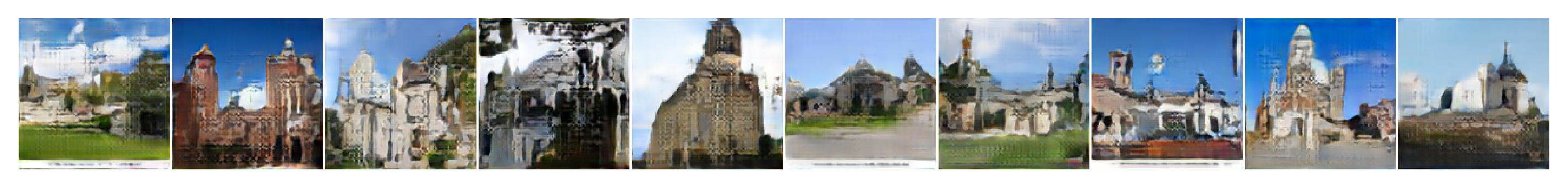}
\caption{Examples of images reconstructed and generated by considered generative models.
All CelebA and LSUN images in this figure and other images produced by PIONEER in the figures below have resolution 128$\times$128.}
\label{fig:modeldemo}
\end{figure}

\begin{figure}[h!]
\pertheader
\centering
\includegraphics[width=1.0\textwidth,trim={0.25cm 0.25cm 0.25cm 0.25cm},clip]{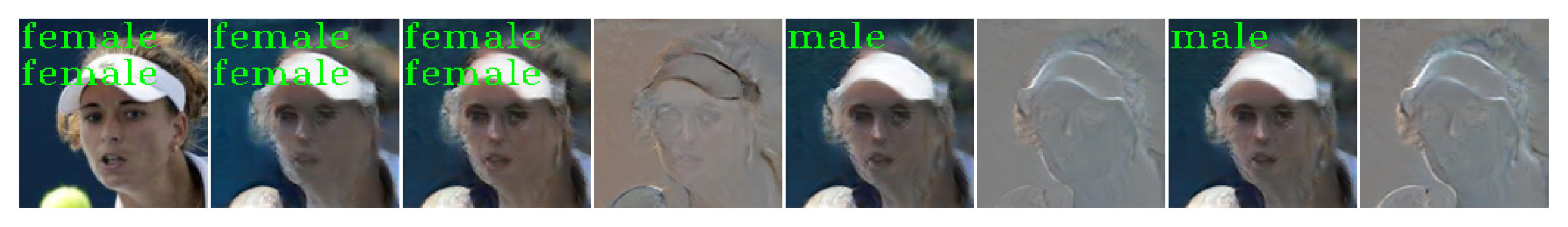}\vspace{-0.1cm}
\includegraphics[width=1.0\textwidth,trim={0.25cm 0.25cm 0.25cm 0.25cm},clip]{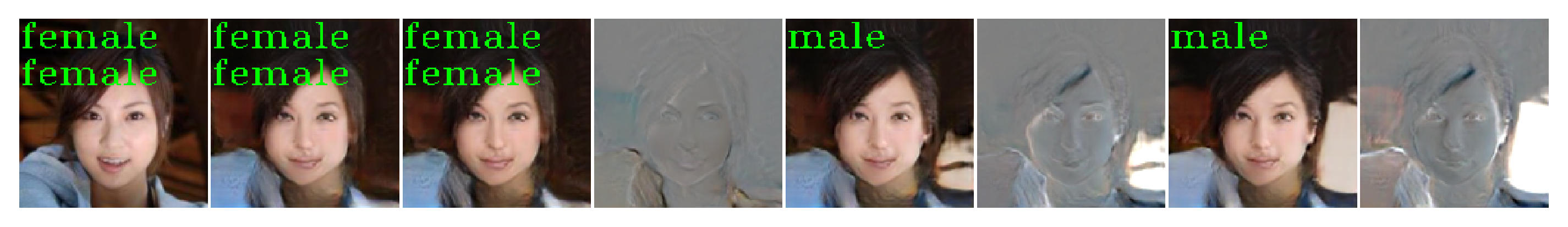}\vspace{-0.1cm}
\includegraphics[width=1.0\textwidth,trim={0.25cm 0.25cm 0.25cm 0.25cm},clip]{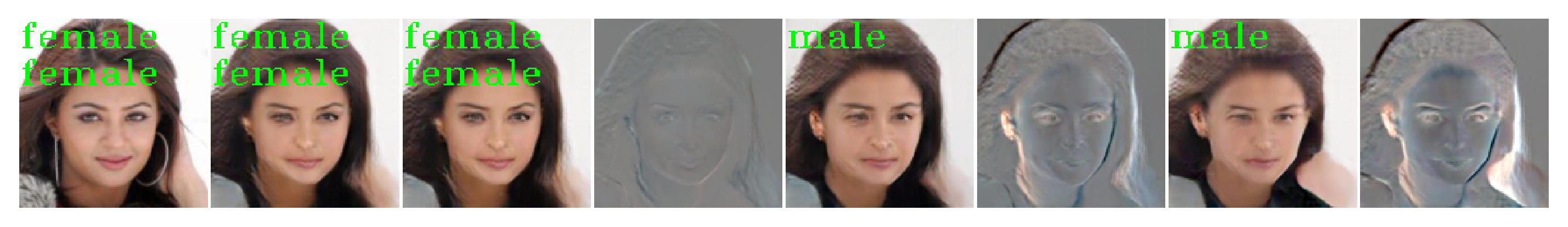}\vspace{-0.1cm}
\includegraphics[width=1.0\textwidth,trim={0.25cm 0.25cm 0.25cm 0.25cm},clip]{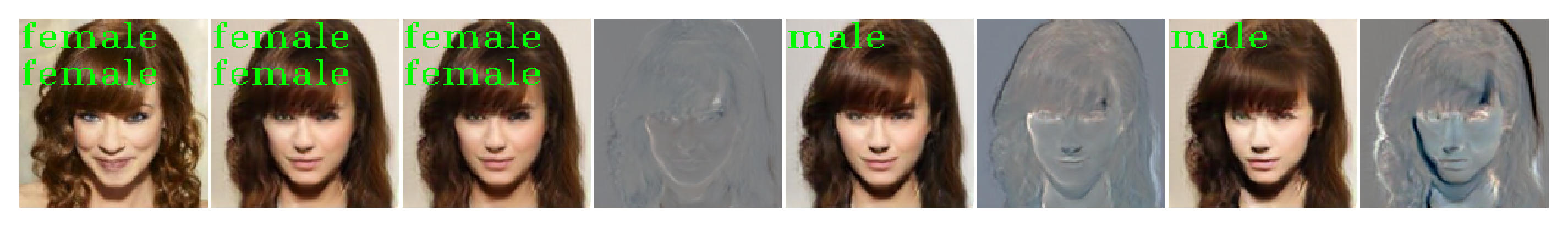}\vspace{-0.1cm}
\includegraphics[width=1.0\textwidth,trim={0.25cm 0.25cm 0.25cm 0.25cm},clip]{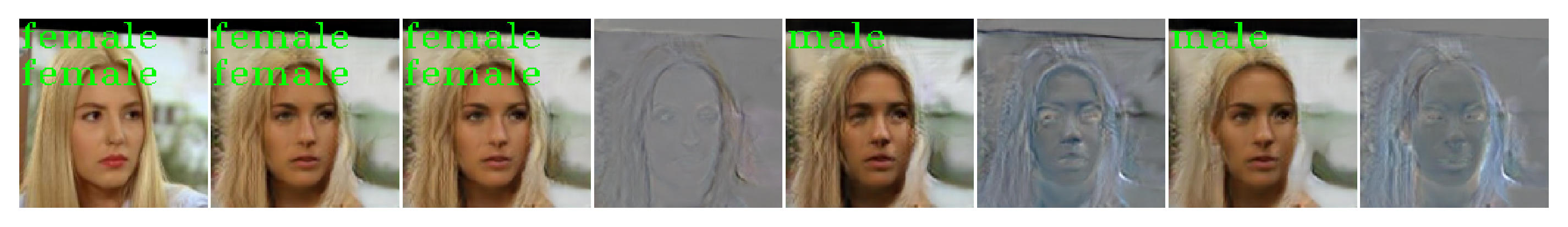}\vspace{0.0cm}
\includegraphics[width=1.0\textwidth,trim={0.25cm 0.25cm 0.25cm 0.25cm},clip]{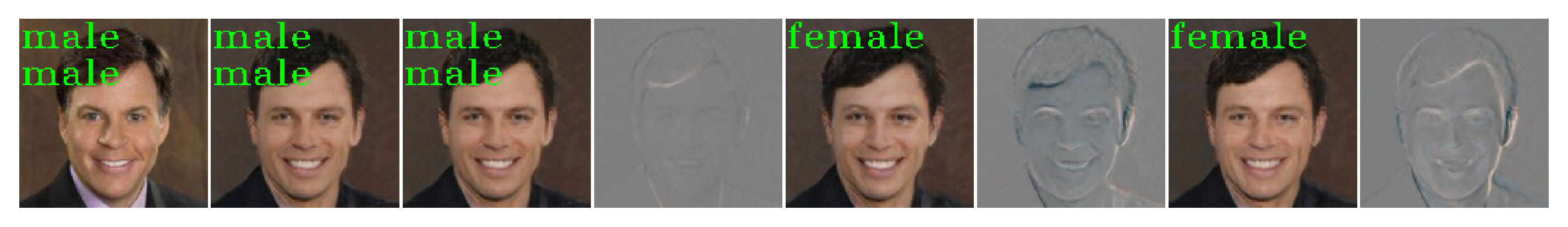}\vspace{-0.1cm}
\includegraphics[width=1.0\textwidth,trim={0.25cm 0.25cm 0.25cm 0.25cm},clip]{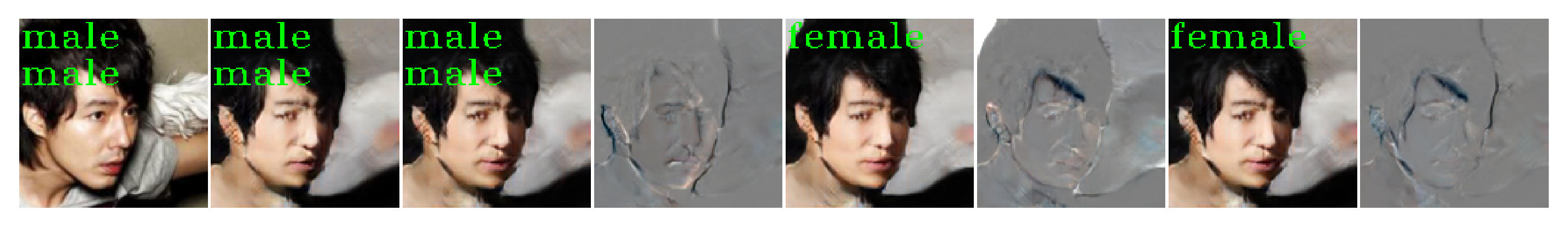}\vspace{-0.1cm}
\includegraphics[width=1.0\textwidth,trim={0.25cm 0.25cm 0.25cm 0.25cm},clip]{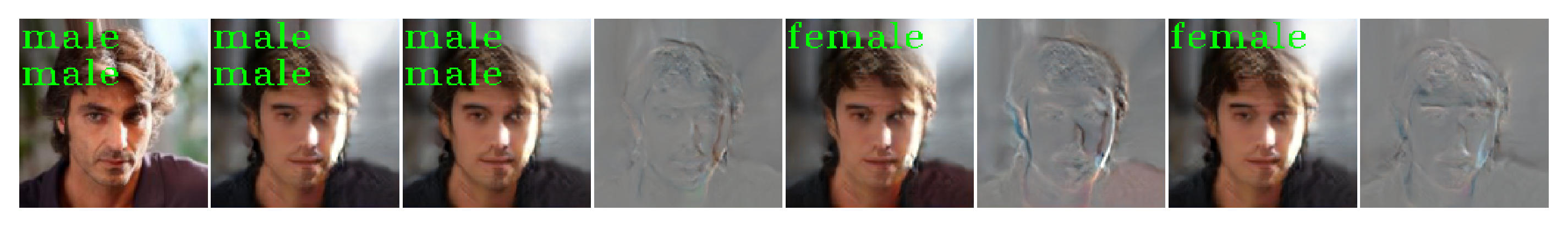}\vspace{-0.1cm}
\includegraphics[width=1.0\textwidth,trim={0.25cm 0.25cm 0.25cm 0.25cm},clip]{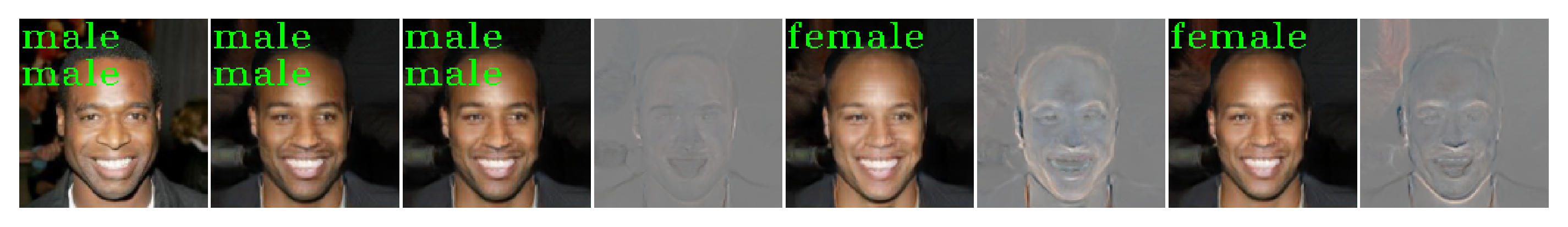}\vspace{-0.1cm}
\includegraphics[width=1.0\textwidth,trim={0.25cm 0.25cm 0.25cm 0.25cm},clip]{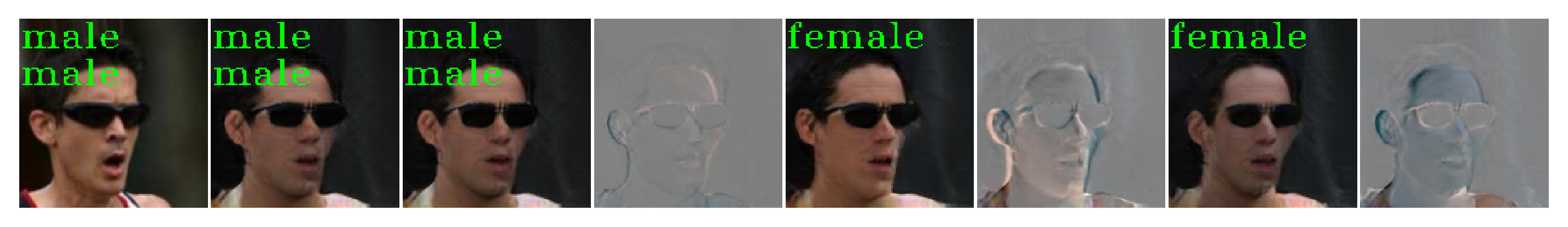}
\caption{Additional examples of approximately minimum latent CelebA image perturbations with $\epsilon = 1~(d = 0.293)$.
Images are arranged as in Fig.~\ref{fig:latent_pert_example_minimum}.
}
\label{fig:more_perturbations_celeba}
\end{figure}

\begin{figure}[h!]
\pertheader
\centering
\includegraphics[width=1.0\textwidth,trim={0.25cm 0.25cm 0.25cm 0.25cm},clip]{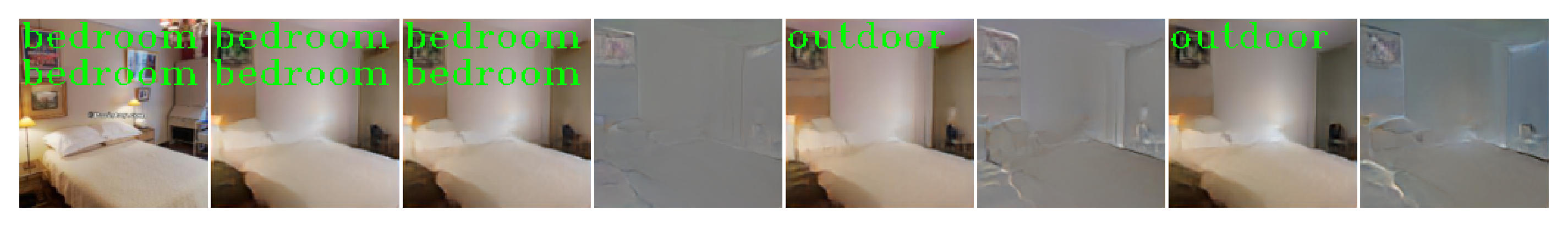}\vspace{-0.1cm}
\includegraphics[width=1.0\textwidth,trim={0.25cm 0.25cm 0.25cm 0.25cm},clip]{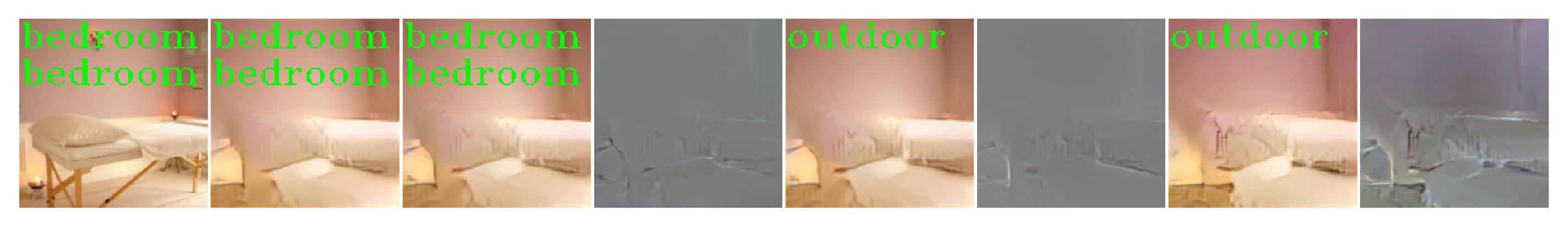}\vspace{-0.1cm}
\includegraphics[width=1.0\textwidth,trim={0.25cm 0.25cm 0.25cm 0.25cm},clip]{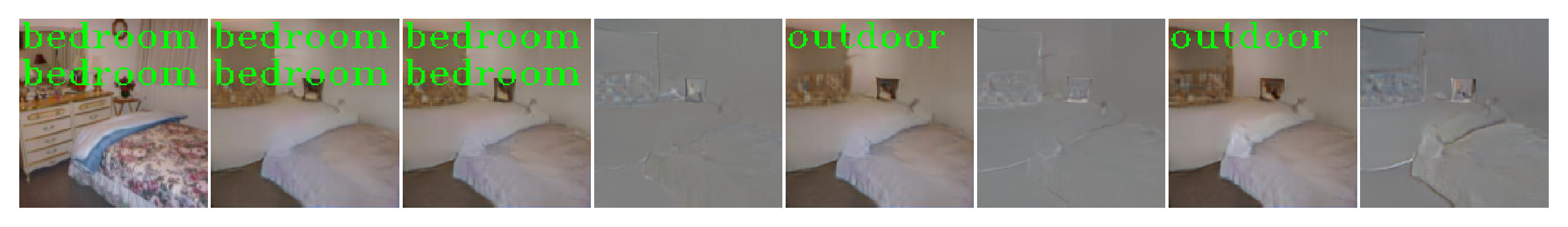}\vspace{-0.1cm}
\includegraphics[width=1.0\textwidth,trim={0.25cm 0.25cm 0.25cm 0.25cm},clip]{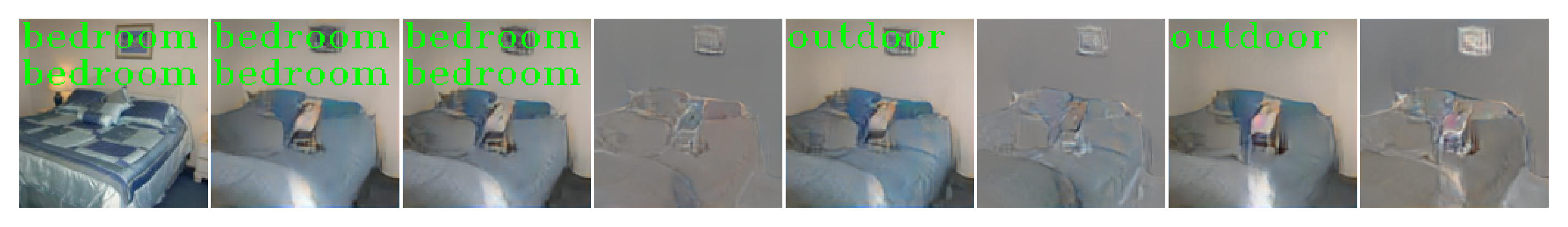}\vspace{-0.1cm}
\includegraphics[width=1.0\textwidth,trim={0.25cm 0.25cm 0.25cm 0.25cm},clip]{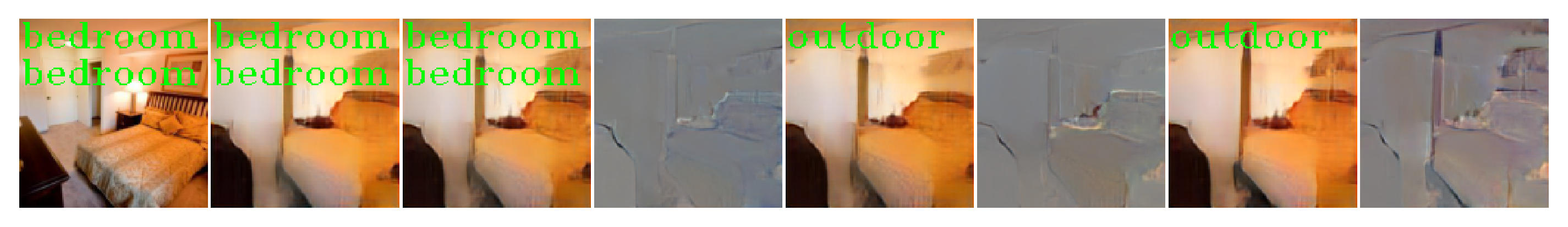}\vspace{0.0cm}
\includegraphics[width=1.0\textwidth,trim={0.25cm 0.25cm 0.25cm 0.25cm},clip]{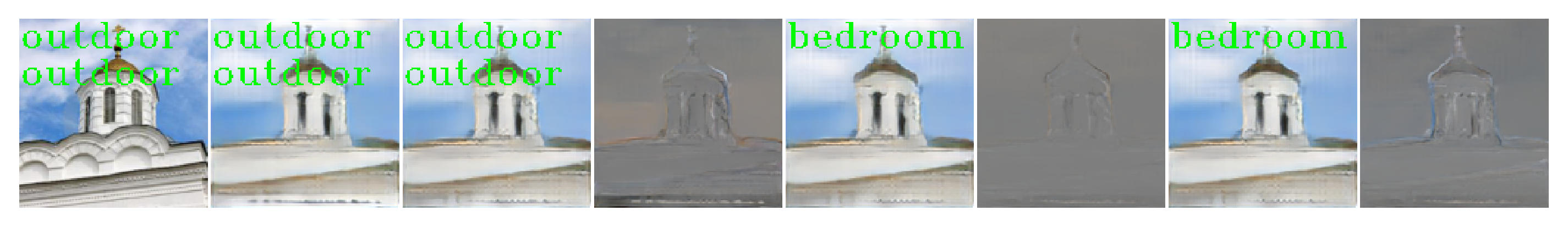}\vspace{-0.1cm}
\includegraphics[width=1.0\textwidth,trim={0.25cm 0.25cm 0.25cm 0.25cm},clip]{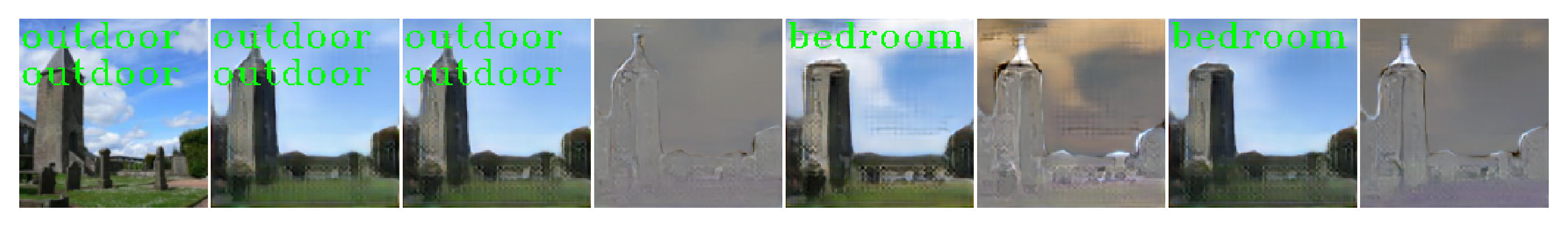}\vspace{-0.1cm}
\includegraphics[width=1.0\textwidth,trim={0.25cm 0.25cm 0.25cm 0.25cm},clip]{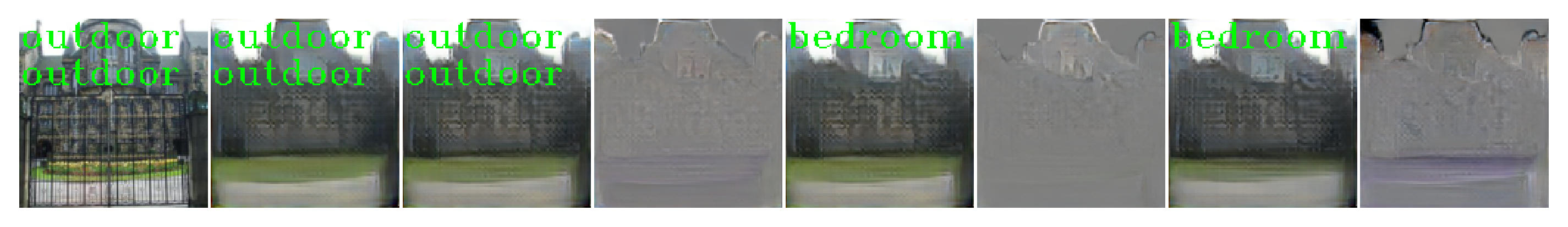}\vspace{-0.1cm}
\includegraphics[width=1.0\textwidth,trim={0.25cm 0.25cm 0.25cm 0.25cm},clip]{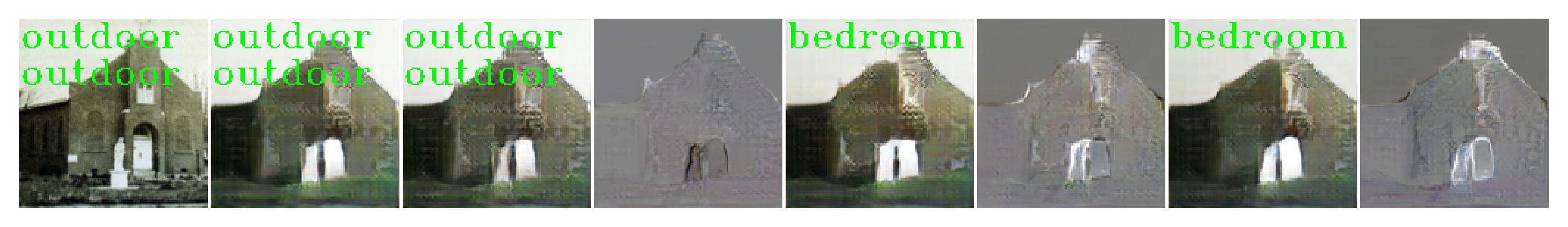}\vspace{-0.1cm}
\includegraphics[width=1.0\textwidth,trim={0.25cm 0.25cm 0.25cm 0.25cm},clip]{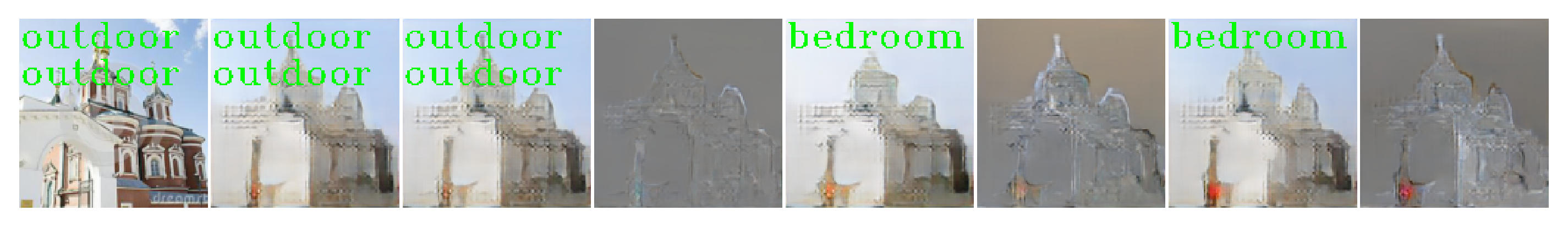}
\caption{Additional examples of approximately minimum latent LSUN image perturbations with $\epsilon = 1~(d = 0.293)$.
Images are arranged as in Fig.~\ref{fig:latent_pert_example_minimum}.
}
\label{fig:more_perturbations_lsun}
\end{figure}

\begin{figure}
\centering
\includegraphics[width=\textwidth]{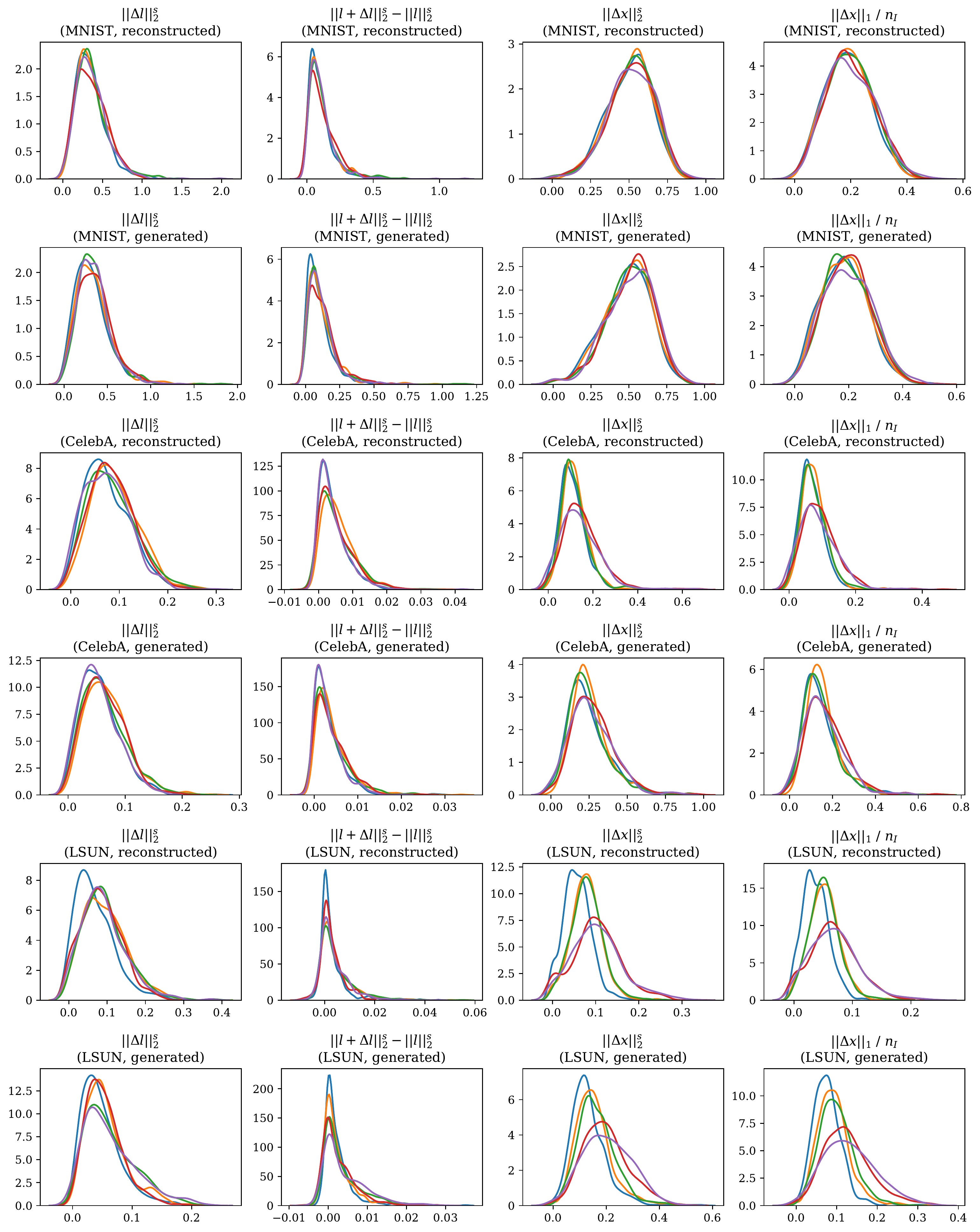}
%MNIST, reconstructed images:
%\includegraphics[width=\textwidth]{{hist-mnist-reconstructed-1.0}.pdf}\vspace{0.1cm}
%MNIST, generated images:
%\includegraphics[width=\textwidth]{{hist-mnist-generated-1.0}.pdf}\vspace{0.1cm}
%CelebA, reconstructed images:
%\includegraphics[width=\textwidth]{{hist-celeba-reconstructed-1.0}.pdf}\vspace{0.1cm}
%CelebA, generated images:
%\includegraphics[width=\textwidth]{{hist-celeba-generated-1.0}.pdf}\vspace{0.1cm}
%LSUN, reconstructed images:
%\includegraphics[width=\textwidth]{{hist-lsun-reconstructed-1.0}.pdf}\vspace{0.1cm}
%LSUN, generated images:
%\includegraphics[width=\textwidth]{{hist-lsun-generated-1.0}.pdf}\vspace{0.1cm}
\caption{
Distribution plots (with Gaussian kernel density estimation) with statistics on approximately minimum latent image perturbations with $\epsilon = 1~(d = 0.293)$ found by PGD (600 images for each row of plots).
$l$ is the decayed latent vector, $\Delta l$ is the found perturbation, and $\Delta x$ is the change of the original image as a vector of pixel intensities.
%Norm scaling is done by dividing the $\ell_2$ norm by $\sqrt{n_I}$ and the $\ell_1$ norm by $n_I$).
Colors correspond to classifiers as follows: \textcolor{blue}{$\mathcal{N}_\text{UT}$ is blue}, \textcolor{orange}{$\mathcal{N}_\text{NR}$ is orange}, \textcolor{darkgreen}{$\mathcal{N}_\text{CA}$ is green}, \textcolor{red}{$\mathcal{N}_\text{R}$ is red}, \textcolor{purple}{$\mathcal{N}_\text{B}$ is purple}.}
\label{fig:distplots}
\end{figure}

\clearpage

\section{Appendix: classifier training procedure}
\label{sec:appendix_training}

All classifiers listed in Section~\ref{sec:setup} were trained as follows.
As the basis for their architecture, we used the script \url{https://github.com/keras-team/keras/blob/master/examples/cifar10_cnn.py}.
Essentially, this is a simple CNN architecture composed of convolutional blocks, ReLU nonlinearities, batch normalization, max-pooling, dropout, and a fully connected layer with $\softmax$ on top.
Training was done with RMSProp.
In each epoch, we took 100 thousand random images from the training set.
The learning rate was set to 0.0004 and multiplied by 0.75 after each epoch.
Training continued for up to 8 epochs, but was stopped prematurely if validation accuracy had not increased during the previous epoch.
The following was specific to different classifier types:
\begin{enumerate}
\item $\mathcal{N}_\text{UT}$: No data augmentation was used. Training was stopped after one epoch.
\item $\mathcal{N}_\text{NR}$: No data augmentation was used. Training was done for the remaining 7 epochs starting from $\mathcal{N}_\text{UT}$.
\item $\mathcal{N}_\text{CA}$: Training images were augmented with conventional approaches: small affine transformations, color distortions and erasures of small image parts.
\item $\mathcal{N}_\text{R}$: Training images were augmented with Gaussian noise of magnitude $\sigma = 0.8$ (pixel intensities belong to $[-1, 1]$).
Note that this is different from the work~\cite{ford2019adversarial}, where for each image first $\sigma$ was selected uniformly at random and then noise was added.
Training was started from $\mathcal{N}_\text{NR}$.
\item $\mathcal{N}_\text{B}$: Training images were first augmented conventionally (as in the case of $\mathcal{N}_\text{CA}$), and then with Gaussian noise (as in the case of $\mathcal{N}_\text{R}$).
Training was started from $\mathcal{N}_\text{CA}$.
\end{enumerate}

%We also redefine the terms \emph{latent vector} to refer to a vector transformed with $t$ and \emph{latent distribution} $\mathcal{D}_L^{i}$ to refer to a distribution of these vectors ($N(0, I)$).
%The aforementioned technique ignores potential dependencies between the components of vectors drawn from $\mathcal{D}_L^{' c=i}$.


\end{document}